\documentclass[runningheads]{llncs}

\usepackage{eccv}

\usepackage{eccvabbrv}

\usepackage{graphicx}
\usepackage{booktabs}
\usepackage{makecell}

\usepackage{multirow}
\usepackage{array}
\usepackage{dsfont}

\usepackage[normalem]{ulem}

\usepackage[accsupp]{axessibility}  

\usepackage{graphicx}
\usepackage{subcaption}

\usepackage[ruled,vlined,linesnumbered]{algorithm2e}
\usepackage{amsmath,amssymb}
\usepackage{setspace}

\usepackage{hyperref}

\usepackage{orcidlink}

\begin{document}

\title{Capacity-Controlled Multi-View Stylization of 3D Gaussian Splatting}

\titlerunning{SceneStyler}

\author{
Zhihao Wen\inst{1} \and
Yixin Yang\inst{1} \and Bojian Wu \inst{2}\orcidlink{0009-0007-1945-8707} \and Yang Zhou\inst{1}\thanks{Corresponding author}\orcidlink{0000-0002-8921-312X}
\and \\Dani Lischinski\inst{3}\orcidlink{0000-0002-6191-0361} \and  Daniel Cohen-Or\inst{1,4}\orcidlink{0000-0001-6777-7445} \and Hui Huang\inst{1}\orcidlink{0000-0003-3212-0544}
}

\authorrunning{Wen et al.}

\institute{Guangdong Provincial Key Laboratory of Visual Media and Multidimensional Intelligence, CSSE, Shenzhen University, China \and Tencent Games, China
\and The Hebrew University of Jerusalem, Israel 
\and Tel Aviv University, Israel \\
\email{\textcolor{magenta}{https://vcc2310.github.io/SceneStyler/}}} 

\maketitle


\begin{abstract}
While 3D Gaussian Splatting (3DGS) provides an efficient and explicit representation for novel view synthesis, enforcing stylistic coherence across viewpoints remains challenging. Existing 3D stylization methods typically apply 2D feature-matching losses independently per rendered view, which leads to unstable style allocation, many-to-one feature reuse, and limited cross-view consistency.
We propose a capacity-controlled framework for multi-view stylization of 3DGS, grounded in optimal transport. Specifically, we reformulate local style matching as a semi-balanced optimal transport problem. 
By introducing explicit column-capacity constraints with tunable strength, our formulation mitigates many-to-one matching and enables controllable allocation of style features. This transport-based objective provides a principled mechanism for balancing feature coverage and stylistic diversity while maintaining stable correspondences across viewpoints.
To further enhance cross-view coherence, we incorporate a novel cross-view matching guidance to constrain correspondences between scene content and style patterns. In addition, we introduce several geometric regularizations to enhance the vanilla 3DGS, thereby enabling optimized Gaussian primitives to represent finer-grained textures during stylization.
Extensive experiments demonstrate that our approach significantly improves multi-view stylistic consistency and produces stable, expressive 3D stylizations while preserving the core semantic structure of the scene. 
\keywords{3D Gaussian Splatting \and Style Transfer \and Optimal Transport}
\end{abstract}
\section{Introduction}
\label{sec:introduction}
The advent of 3D Gaussian Splatting (3DGS)~\cite{3dgs} has marked a significant breakthrough in novel view synthesis, enabling real-time photorealistic rendering with exceptional quality. 
The explicit and structured representation of 3DGS 
also enables intuitive appearance editing of scenes~\cite{stylegaussian,stylizedgs}. This capability naturally extends to 3D stylization: transferring the artistic style of a 2D reference image to a 3D scene while preserving its content structure. However, ensuring \emph{detailed and consistent stylization across multiple viewpoints} remains a fundamental challenge, limiting the practical deployment of such technology in augmented reality, virtual production, and immersive digital art.

Existing 3DGS stylization methods typically apply 2D style losses independently to each rendered view~\cite{stylizedgs, sgsst, abc-gs, clipgaussian, stylegaussian}. While effective at transferring local texture details, this view-independent paradigm suffers from severe \emph{multi-view inconsistency} for two main reasons: First, \emph{within each individual view}, standard feature matching strategies such as Nearest Neighbor Feature Matching (NNFM)~\cite{arf} suffer from the \emph{many-to-one matching problem}: multiple content features are matched to the same dominant style feature, resulting in repetitive local textures and limited style diversity. Second, \emph{across different views}, the same 3D scene content projected to different viewpoints may be matched to \emph{different} style patterns due to viewpoint-dependent feature variations, causing blurred or inconsistent stylization when observed from novel angles.


Recent efforts to address multi-view consistency have explored using DINO features~\cite{multistylegs} or geometric constraints~\cite{stylegaussian}. However, these approaches only enforce consistency \emph{alongside} the matching process---either by augmenting image features with global descriptors or regularizing geometry---without addressing the \emph{matching mechanism itself}. Consequently, they neither prevent many-to-one matching within individual views, nor explicitly constrain the same 3D content to match consistent style patterns across viewpoints.

In this work, we propose a principled framework that tackles both root causes of multi-view inconsistency. First, we introduce a \textbf{Capacity-Controlled Feature Transport (CCFT) loss} based on \textbf{semi-balanced optimal transport}. By formulating style matching as an optimal transport problem with explicit \textbf{column-capacity constraints}, our method \emph{systematically guarantees} the suppression of many-to-one matching within each view, while providing \emph{tunable control} over style pattern diversity via the capacity parameter. Second, to address cross-view misalignment, we propose a \textbf{Cross-View Matching Guidance} that regularizes feature correspondences across adjacent viewpoints, explicitly constraining the same 3D content to match consistent style patterns regardless of viewpoint variations.
To support fine-grained texture representation, we additionally introduce a set of geometric constraints into vanilla 3DGS to yield smaller, more uniform Gaussian primitives, establishing a robust, finer geometric foundation for high-quality stylization.

In summary, our main contributions are:
\begin{itemize}
    \item We introduce a novel Capacity-Controlled Feature Transport (CCFT) loss based on semi-balanced optimal transport. Our formulation suppresses many-to-one matching via explicit column-capacity constraints, with a tunable strength that controls the diversity of style patterns.
    \item We propose Cross-View Matching Guidance to enhance multi-view consistency, explicitly encouraging adjacent viewpoints to match the same 3D content to consistent style patterns.
    \item We develop an enhanced scene reconstruction procedure that produces smaller, more uniform Gaussians to represent fine-grained surface textures, providing a robust foundation for high-quality stylization.
    \item Extensive experiments demonstrate superior visual quality and multi-view consistency of our approach compared to state-of-the-art methods.
\end{itemize}

\section{Related Work}
\label{sec:related_work}


\paragraph{Neural Style Transfer.} 
Creating stylized/artistic content is a hot topic in vision and graphics. Since the seminal work of Gatys et al.~\cite{nst}, significant progress has been made in neural style transfer for images. 
Existing work mainly focuses on how to effectively model image content and represent style. Beyond the classical statistic-based Gram loss~\cite{nst}, many successful style losses have been proposed, including histogram loss~\cite{histogram}, Wasserstein loss~\cite{strotss,Heitz_2021_CVPR}, nearest neighbor matching loss~\cite{cnnmrf, nts}, GAN loss~\cite{DoesFS24}, token-based loss on DINO-ViT~\cite{splicing}, and attention features-based losses on diffusion networks~\cite{AttDistill25}.
Despite these advances, directly applying such 2D methods to 3D stylization remains challenging, particularly in maintaining multi-view consistency and preserving scene geometry.


\paragraph{3D Style Transfer.} 

The advent of Neural Radiance Field (NeRF)~\cite{nerf} has made neural style transfer feasible for 3D scenes. Leveraging this representation~\cite{nerf, tensorrf, plenoxels, nerf++, instantngp}, NeRF-based 3D stylization has been explored through both iterative~\cite{snerf, stylizednerf, arf, arfplus, tcstyle} and feed-forward methods~\cite{hypernetwork, stylerf, styledyrf, fprf}.
Among these works, ARF~\cite{arf} achieves high-quality and view-consistent stylization by introducing a Nearest-Neighbor Feature Matching (NNFM) loss. However, NeRF-based methods are generally limited by their lengthy training and rendering costs.

Recently, 3DGS~\cite{3dgs} 
has rapidly emerged as a cornerstone in 3D vision thanks to its outstanding rendering efficiency and quality, paving the way for 3DGS-based style transfer.
Some works follow the feature statistics paradigm: StyleGaussian~\cite{stylegaussian} builds upon AdaIN~\cite{adain}, incorporating feature embedding and a 3D decoder to achieve feed-forward stylization, yet struggles to capture fine-grained patterns, resulting in limited style quality. SGSST~\cite{sgsst} introduces a multiscale style loss based on the Gram matrix~\cite{nst}, but the global statistics it uses often lead to content-style mismatch. Other works adapt the nearest-neighbor feature matching to 3DGS: StylizedGS~\cite{stylizedgs} augment the NNFM loss~\cite{arf} with additional guidance to enhance user controllability; G-Style~\cite{gstyle} incorporates low-frequency features and a densification strategy to improve the expression of style patterns. Nonetheless, these methods tend to produce monotonous styles due to the inherent ``many-to-one'' nature of NNFM. ABC-GS~\cite{abc-gs} mitigates this shortcoming via feature alignment, but also over-smooths high-frequency features. CLIPGaussian~\cite{clipgaussian} takes a different approach by leveraging CLIP model~\cite{clip} for multi-modal style guidance. While demonstrating strong generalization, the CLIP-based method falls short in style fidelity compared to VGG feature-based methods on image style references.

It is worth noting that, all the aforementioned 3DGS stylization methods ignored style coherence across viewpoints. MultiStyleGS~\cite{multistylegs} augmented image features with DINO feature, aiming to improve the uniqueness of rendered features. StylizedGS~\cite{stylizedgs} proposed a depth preservation loss to enhance geometric consistency. Nonetheless, there is still no work that directly improves multi-view consistency within the feature-matching mechanism.

In this work, we focus on addressing the primary obstacles affecting coherent 3DGS stylization. Our method improves reconstruction quality for better stylization, solves the many-to-one matching issue, and addresses multi-view consistency mechanistically, finally producing superior stylized results. 

\begin{figure*}[t]
\centering
\includegraphics[width=\linewidth]{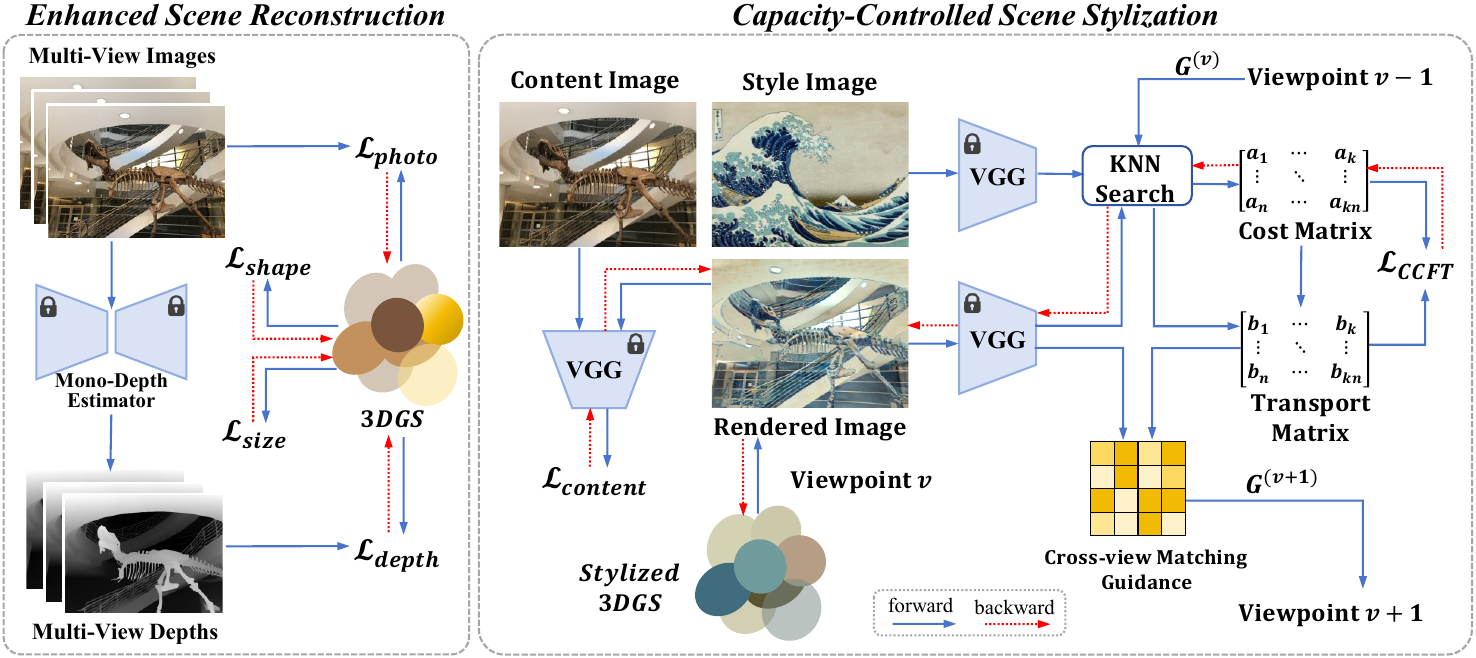}
\caption{\textbf{Method Overview}. Our framework contains two stages: enhanced reconstruction and capacity-controlled stylization. In the reconstruction stage, we optimize the 3D Gaussians using estimated depth as geometric constraints to achieve more accurate scene reconstruction. Meanwhile, we regularize the scale and shape of the primitives to better represent stylized textures later. Then, in the stylization stage, we introduce a Capacity-Controlled Feature Transport (CCFT) loss and cross-view matching guidance to obtain coherent stylized views. With a content loss that preserves scene structure, we update the color parameters of 3D Gaussians, yielding the final stylized 3D scene.} 
\label{fig:pipeline}
\end{figure*}

\section{Method}
\label{sec:method}


Given a set of multi-view images, our goal is to reconstruct the scene with 3DGS~\cite{3dgs} and transfer the scene's appearance into the style of a user-provided reference image. As depicted in Fig.~\ref{fig:pipeline}, we first train the 3DGS with additional geometric regularizers to get an enhanced scene representation; then, we stylize the Gaussian primitives view-by-view using a novel optimal-transport-based style loss, which addresses the many-to-one matching problem and multi-view consistency, yielding high-quality, coherent scene stylization.


\subsection{Preliminary of Gaussian Splatting}
3DGS~\cite{3dgs} models a scene explicitly as a collection of anisotropic 3D Gaussian primitives. Each primitive is defined by a set of parameters: a center position $\mu_i$, a covariance matrix $\Sigma_i$, an opacity $\alpha_i$, and view-dependent color $\mathbf{c}_i$ represented using spherical harmonics coefficients. 
During rendering, each 3D Gaussian is projected into screen space through an affine transformation. The color of a pixel is then computed by blending the contributions of all Gaussians that overlap the pixel, ordered by their depth, using the alpha blending formula:

\begin{equation}
P_{pixel}=\sum_{i} \mathbf{c}_i \alpha_i \prod_{j=1}^{i-1}(1-\alpha_j)
\end{equation}

To achieve high performance, 3DGS employs a tile-based rasterization coupled with a CUDA-accelerated differentiable renderer, enabling rapid image synthesis and gradient computation while balancing high-quality rendering. 


\subsection{Enhanced Scene Reconstruction}
While vanilla 3DGS can faithfully synthesize novel views of the original scene, its highly anisotropic nature often leads to irregular distributions and deformations of individual Gaussians, which can severely impact stylization quality. StylizedGS~\cite{stylizedgs} and ABC-GS~\cite{abc-gs} identify a similar issue, and both propose periodically filtering out Gaussians with low opacity or large scale before the stylization stage. However, they didn't choose to holistically correct the scene geometry. In contrast, we introduce a series of regularization terms targeting geometric parameters during the reconstruction stage. Coupled with the density control mechanism, we perform an enhanced reconstruction to improve overall scene geometry and regularize the shapes of individual 3D Gaussians, facilitating stable, high-quality stylization of the next step.

First, we employ a depth loss to enhance the spatial accuracy of 3DGS, encouraging each Gaussian to concentrate on the valid objects, mitigating the floaters within the scene. Specifically, we compute the L1 distance between the rendered depth and the estimated depth:
\begin{equation}
\mathcal{L}_{\mathit{depth}} = \left\| D_{\mathit{render}} - D_{\mathit{pred}} \right\|_1,
\text{where}\quad D_{\mathit{render}} = \sum_{i} d_i \alpha_i \prod_{j=1}^{i-1}(1-\alpha_j)
\end{equation}
Here $d_i$ is the depth of the $i$-th Gaussian primitive, and the predicted depth $D_{\mathit{pred}}$ is acquired from the training images using a pretrained depth estimator~\cite{dav2}.

Beyond the depth, the shape and scale of Gaussian primitives directly affect the quality of stylization. To efficiently represent high-frequency textures in real views, vanilla 3DGS typically optimizes Gaussians into highly irregular shapes. However, extremely elongated primitives can introduce geometric artifacts in stylized views. We, instead, introduce a shape regularization term:
\begin{equation}
\mathcal{L}_{\mathit{shape}} = \frac{1}{\mathit{N_{gs}}}\sum_{i=1}^{\mathit{N_{gs}}}
\left[ S_{\mathit{max}}^i - \lambda_s S_{\mathit{min}}^i \right]_+, 
\text{where}\quad [x]_+ = \max\{x, 0\}
\end{equation}
where $\mathit{N_{gs}}$ is the total number of Gaussian primitives, $S_{\mathit{max}}^i$ and $S_{\mathit{min}}^i$ denote the longest and the shortest axis of the $\mathit{i}$-th Gaussian primitive, respectively. The hyperparameter $\lambda_s$ represents the expected ratio between the longest and the shortest axis. With this regularization, the shape uniformity of Gaussian primitives can be fairly enforced.

Furthermore, the original 3DGS tends to produce large-scale Gaussian primitives to efficiently represent regions with fewer textures in real views. However, large-scale primitives typically occupy large spaces and have low density, making it difficult to express detailed style features. Therefore, we also introduce a scale regularization term:
\begin{equation}
\mathcal{L}_{\mathit{size}} = \frac{1}{\mathit{N_{gs}}}\sum_{i=1}^{\mathit{N_{gs}}} 
\left( \left[ S_{\mathit{max}}^i - \beta_1 \right]_+ 
+ \left[ \beta_2 - S_{\mathit{max}}^i \right]_+ \right)
\end{equation}
where $\beta_1$ and $\beta_2$ are the hyperparameters that represent the upper and lower bounds of the longest axis, respectively. This regularization can normalize large-scale Gaussian primitives to smaller ones within a bounded range, thereby capturing stylistic details.

The total loss function used for our enhanced scene reconstruction is:
\begin{equation}
\mathcal{L}_{\mathit{rec}} = \mathcal{L}_{\mathit{photo}} + \lambda_{\mathit{depth}} \mathcal{L}_{\mathit{depth}} + \lambda_{\mathit{shape}} \mathcal{L}_{\mathit{shape}} + \lambda_{\mathit{size}} \mathcal{L}_{\mathit{size}}
\end{equation}
where $\mathcal{L}_{\mathit{photo}}$ is the original training loss of 3DGS, \ie, the linear combination of L1 loss and SSIM loss between the rendered image and the real image.

\begin{figure}[t]
    \centering
    \includegraphics[width=\linewidth]{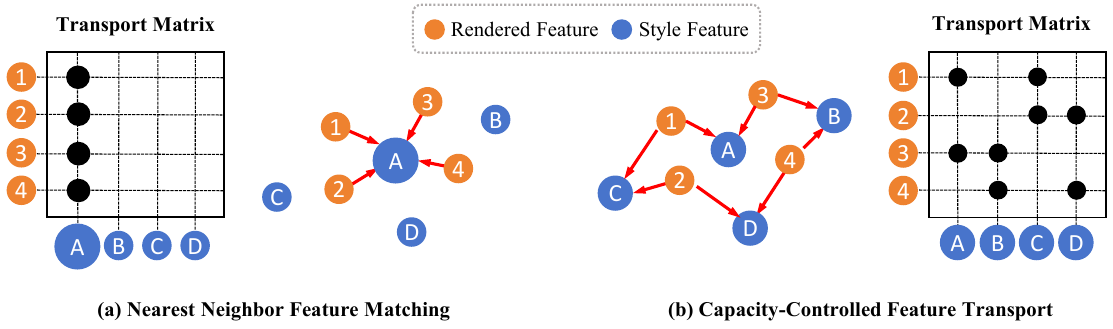}
    \caption{\textbf{Capacity-Controlled Feature Transport}. The capacity of each style feature is represented by the size of its node, with larger nodes indicating greater capacity. (a) When there's no capacity constraint, multiple rendering features are mapped to the same style feature, and matching degrades to nearest-neighbor search, resulting in the many-to-one issue. (b) With capacity control, the matching of each style feature tends to be equalized, and more style features are utilized.}
    \label{fig:CFT_loss}
\end{figure}

\subsection{Capacity-Controlled Scene Stylization}
As discussed previously, while the NNFM loss~\cite{arf} produces more visually appealing results in multi-view stylization, it still suffers from overuse of a limited subset of style features, i.e., the many-to-one issue, and it completely ignores multi-view consistency. To address these issues, we first reformulate the nearest neighbor matching problem as a local feature assignment task solved via optimal transport, and then introduce a novel cross-view guidance to further enhance matching coherence across different viewpoints. 

\subsubsection{Capacity-Controlled Feature Transport.} 
To prevent many-to-one matching, we formulate style transfer as a capacity-constrained optimal transport problem, where each style feature has a tunable capacity limit restricting the number of matched rendered features. This gives rise to our Capacity-Controlled Feature Transport (CCFT) loss (Fig.~\ref{fig:CFT_loss}), which inherently prevents degenerate matching while enabling flexible control over style diversity via the capacity parameter.

Formally, let $X=\{x_i\}_{i=1}^N$ denote the set of rendered image features and $Y=\{y_j\}_{j=1}^M$ denote the set of style image features. For each rendered feature, we search for K nearest neighbor style features using cosine similarity, with indices denoted as $\mathrm{KNN}(i)$, which forms a bipartite sparse support:
\begin{equation}
E = \{ (i, j) \mid j \in \mathrm{KNN}(i) \}
\end{equation}
We then solve an entropy-regularized semi-balanced optimal transport problem on the bipartite sparse support:
\begin{equation}
\begin{gathered}
T = \arg\min_{T \geq 0} \left[ \langle T, C \rangle_E - \varepsilon H(T)_E + \tau \mathrm{KL}(T^\top \overrightarrow{1} \| b) \right] \\
\text{s.t.}\quad T \overrightarrow{1} = a, \quad a_i = \frac{1}{N}
\end{gathered}
\end{equation}
Here, $\mathit{T}$ is the transport matrix to be optimized, and $\mathit{C}$ is the cost matrix representing the pairwise distances between rendered and style features. $H(T)$ is the entropy regularization term, with parameter $\varepsilon$ controls its strength. 
$a$ is the row capacity constraint, set to a uniform distribution. $b$ is the column capacity constraint, defined as a uniform distribution over the reachable column set:
\begin{equation}
b_j = 
\begin{cases} 
\frac{1}{|J|}, & j \in J \\
0, & j \notin J
\end{cases} 
\quad \text{where}\quad J = \{ j : \exists i, (i, j) \in E \}
\end{equation}
The column capacity constraint is softly enforced using KL divergence, with parameter $\tau$ controlling the degree of uniform allocation among style features. After iteratively solving for the optimal transport matrix $\mathit{T}$ using the Sinkhorn-Knopp algorithm~\cite{cuturi2013sinkhorn}, our CCFT loss is finally computed as the transport cost under the optimal plan:
\begin{equation}
\mathcal{L}_{\mathit{CCFT}} = \langle T, C \rangle_E = \sum_{(i,j) \in E} T_{ij} C_{ij}
\end{equation}

\subsubsection{Cross-view Matching Guidance.} 
Ensuring multi-view consistency is crucial for high-quality stylization. Although the CCFT loss enables smooth, controllable feature assignment at each iteration, content changes across viewpoints can lead to unstable bipartite sparse support when searching for nearest-neighbor solely based on the similarity between rendered and style features, thereby compromising multi-view consistency. Inspired by Guided Correspondence Distance~\cite{nts}, we incorporate a cross-view matching guidance in the 
distance computation of nearest neighbor searching:
\begin{equation}
\begin{gathered}
\text{Distance}({x}_i^{(v)},{y}_{j}) = \text{Sim}({x}_i^{(v)},{y}_{j})+\lambda_{guide}\,\text{Sim}({x}_i^{(v)},{g}_{j}^{(v)})
\\
\text{where}\quad {g}_{j}^{(v)}=\sum_{i}T_{ij}^{(v-1)}\,{x}_{i}^{(v-1)}
\end{gathered}
\end{equation}
\begin{figure}[t]
    \centering
    \includegraphics[width=\linewidth]{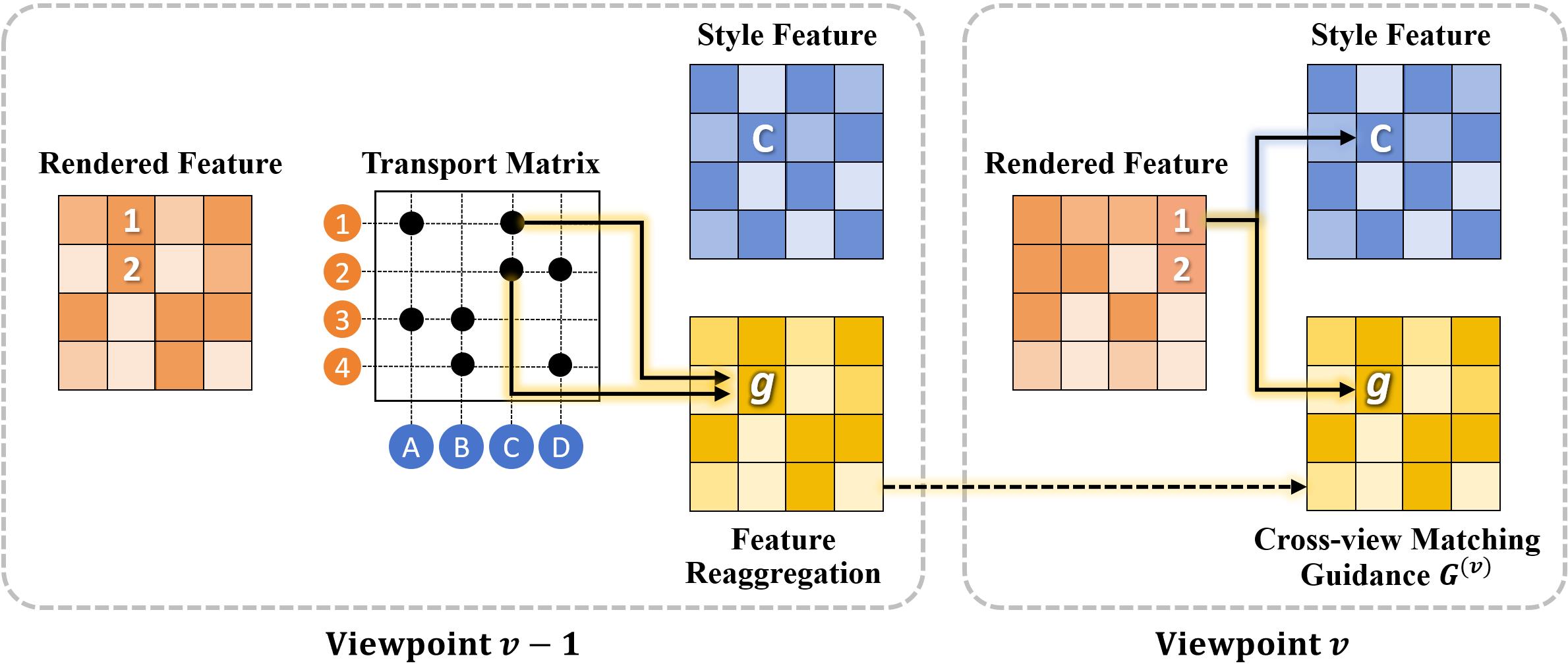}
    \caption{\textbf{Cross-view Matching Guidance}. We introduce a guidance map to improve matching coherence across viewpoints by reaggregating rendered features according to the transport matrix. Here, for example, at viewpoint $v-1$, suppose rendered features \texttt{1}$^{(v-1)}$ \& \texttt{2}$^{(v-1)}$ are assigned to style feature \texttt{C} in the optimal transport. Then, the guidance feature \texttt{\textit{g}}$^{(v)}$ for \texttt{C} is aggregated from \texttt{1}$^{(v-1)}$ \& \texttt{2}$^{(v-1)}$ according to the transport plan. Finally, at viewpoint $v$, the guided distance between \texttt{1}$^{(v)}$ and \texttt{C} is defined as the weighted sum of the distance between \texttt{1}$^{(v)}$ \& \texttt{C} and the distance between \texttt{1}$^{(v)}$ \& \texttt{\textit{g}}$^{(v)}$.}
    \label{fig:guidance}
\end{figure}
Here Sim($\cdot$,$\cdot$) represents cosine distance, and the superscript $(v)$ denotes the iteration at viewpoint $v$. The modified/guided distance metric 
jointly considers the distance between rendered and style features, and the distance of rendered features to a guidance map computed from the previous iteration. As illustrated in Fig.~\ref{fig:guidance}, the guidance map for viewpoint $v$ is the weighted projection of the rendered features from viewpoint $v-1$ (\ie, the viewpoint of previous iteration) onto the style features according to the optimal transport matrix. The element of the guidance map indicates which rendered features are matched to the corresponding style feature at viewpoint $v-1$. This guidance encourages similar rendered features across different viewpoints to match the same corresponding style feature, thereby stabilizing sparse bipartite support throughout optimization and improving multi-view consistency.

\subsubsection{Content Loss.}
For content preservation of the 3D scene, we calculate the MSE distance between rendered features $X=\{x_i\}_{i=1}^N$ and content features $\hat{X}=\{\hat{x}_{i}\}_{i=1}^N$ as the content loss, which has the following form:
\begin{equation}
\mathcal{L}_{\mathit{content}}=\frac{1}{N}\sum_{i=1}^{N}\left(x_i-\hat{x}_i\right)^2
\end{equation}

Finally, the total loss used for stylization is a weighted combination of the CCFT Loss and the content loss. To further preserve the scene's structure, we fix the geometric attributes of 3DGS during the stylization, optimizing only its color attributes‌. Additionally, we apply color transfer following~\cite{arf} before and after the stylization stage to maintain color consistency with the style image.

\section{Experiments}
\label{sec:experiments}
\subsubsection{Implementation Details.}
Our implementation is based on the original 3DGS and introduces three regularizations during reconstruction, where the parameters $\{\lambda_{s},\beta_1,\beta_2,\lambda_{shape},\lambda_{size}\}$ are set to $\{3.5,0.1,0.001,10,1\}$, and $\lambda_{depth}$ decays exponentially from 1 to 0.01. For stylization, we use a pre-trained VGG-16 network~\cite{vgg} as the feature extractor, and we set parameters $\{K,\varepsilon,\tau,\lambda_{guide},\lambda_{CCFT},\\
\lambda_{content}\}$ to $\{16,0.05,0.5,1,30,0.005\}$. All experiments are performed on a workstation equipped with a single NVIDIA RTX 4090 GPU.

\textbf{Datasets.}
We conduct comprehensive experiments on multiple datasets, including LLFF~\cite{llff}, Tanks and Temples (T\&T)~\cite{tnt}, and Mip-NeRF 360~\cite{mipnerf}, encompassing both forward-facing scenes and unbounded 360$^{\circ}$ environments. To validate robustness across diverse scenarios, we use stylistic images from various artistic categories from the WikiArt dataset~\cite{stylegaussian} as style references.

\textbf{Baselines.}
For comparative analysis, we select state-of-the-art 3DGS-based style transfer methods as baselines, including StylizedGS~\cite{stylizedgs}, ABC-GS~\cite{abc-gs}, SGSST~\cite{sgsst}, CLIPGaussian~\cite{clipgaussian}, and StyleGaussian~\cite{stylegaussian}. With the exception of StyleGaussian, which adopts a feed-forward approach, all comparative methods are optimization-based. For all baselines, we use their officially released code.

\begin{figure}[t]
\centering
\includegraphics[width=\linewidth]{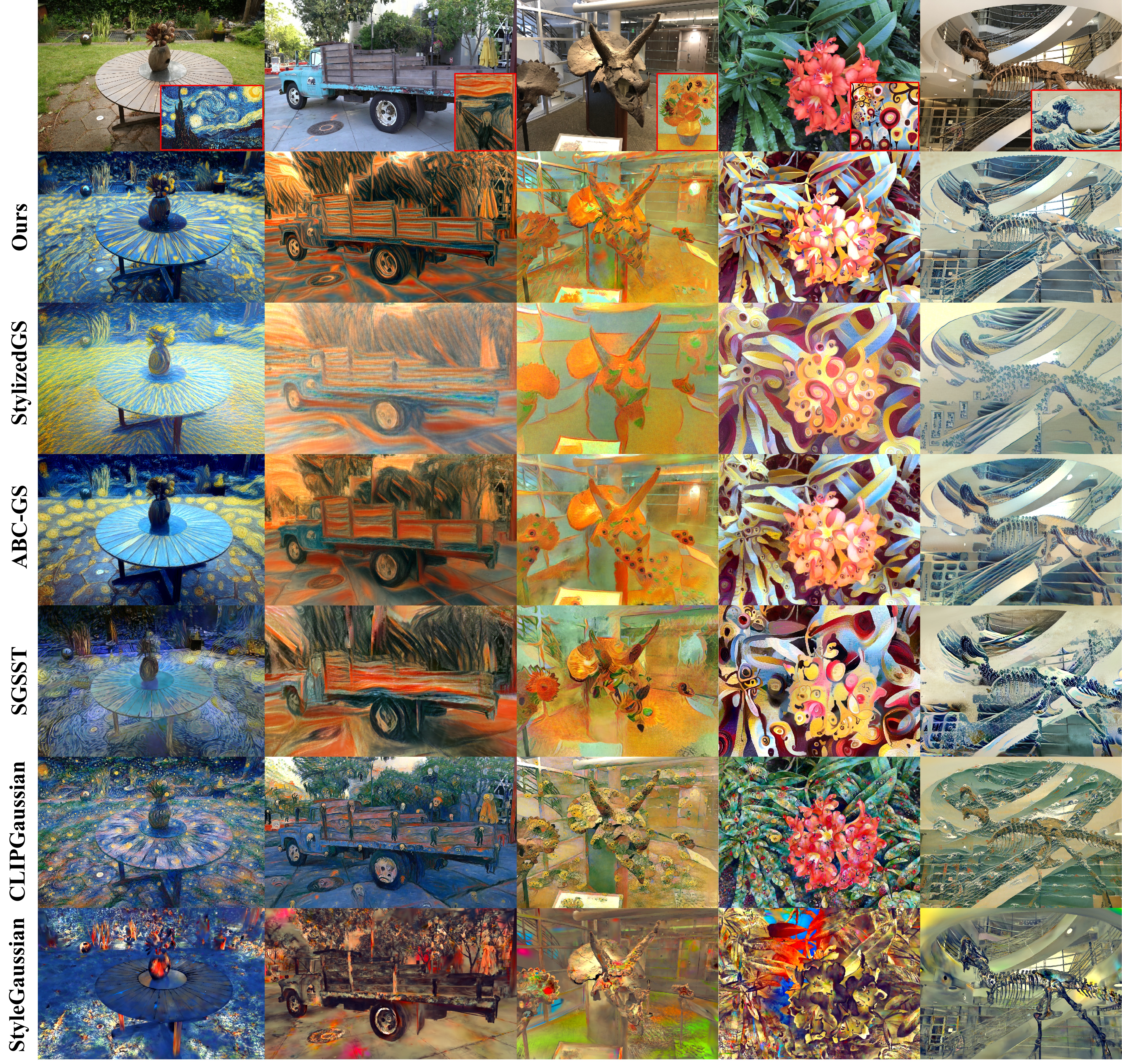}
\caption{\textbf{Qualitative comparison.} Our method demonstrates a significant advantage in reproducing the texture details and brushstrokes in the style references while still preserving the semantic structure of the input scene.}
\label{fig:qualitative}
\end{figure}

\subsection{Experimental Results}

\subsubsection{Qualitative Results.}
Fig.~\ref{fig:qualitative} shows a gallery of visual comparisons between our method and the baselines. StylizedGS employs the NNFM Loss to stylize scenes, where the many-to-one issue leads to a large amount of repetitive textures; for example, numerous yellow stripe-like textures appear (column 1). ABC-GS uses FAST Loss to present more style patterns, but the global linear transformation tends to cause erroneous matches in regions with insufficient local semantics; for example, sunflower seed patterns are incorrectly matched to the triceratops’s mouth (column 3). SGSST achieves a highly global stylization, but when there are significant differences in local structure between content and style, the content structure is severely degraded; for instance, it is difficult to distinguish the shapes of petals and leaves (column 4). CLIPGaussian constructs the style loss using the CLIP model; however, the model exhibits semantic illusions between images, leading to patterns and colors that are inconsistent with the style image. StyleGaussian utilizes AdaIN for instant stylization, but this alignment of low-order statistics makes it difficult to capture the specific style details.

Compared to the baselines, our stylization results demonstrate significant advantages in style matching and content preservation. For example, in the first and second scenes, our method accurately reproduces the painting's brushstrokes; in the last scene, it reproduces texture details, particularly the white droplets in the waves, more evenly than other methods. Meanwhile, our results clearly preserve the semantic content of the input scene; see, e.g., the truck's wheels (column 2) and the staircases' handrails (columns 3 \& 5). These qualitative results provide strong validation of the effectiveness of our stylization strategy.

\begin{table}[t]
\centering
\caption{\textbf{Quantitative comparison}, where ArtFID evaluates style quality, Structure Loss measures content preservation, and the last two metrics indicate multi-view consistency. ``Ours (3DGS)'' denotes the ablated version of our method using the original 3DGS for scene reconstruction. Our full method achieved the best across all metrics.}
\label{tab:quant_eval}
\footnotesize
\setlength{\tabcolsep}{2.5pt} 
\begin{tabular}{lccccc}
\toprule


Method & ArtFID$\downarrow$ & Structure Loss$\downarrow$ & Short MEt3R$\downarrow$ & Long MEt3R$\downarrow$ \\

\midrule
StylizedGS & \underline{23.068} & 0.0462 & 0.1201 & 0.2858 \\
ABC-GS & 25.026 & 0.0357 & \underline{0.1198} & \underline{0.2822} \\
SGSST & 25.143 & 0.0568 & 0.1384 & 0.3064 \\
CLIPGaussian & 27.293 & 0.0431 & 0.1353 & 0.2903 \\
StyleGaussian & 31.855 & 0.0425 & 0.1284 & 0.2982 \\
Ours (3DGS) & {24.133} & \underline{0.0320} & \underline{0.1198} & \textbf{0.2795} \\
Ours & \textbf{22.801} & \textbf{0.0318} & \textbf{0.1196} & \textbf{0.2795} \\

\bottomrule
\end{tabular}
\end{table}






\begin{table}[t]
\centering
\caption{\textbf{User study}. We presented the 6-method results for each scene and asked users to select the top three for each evaluation aspect, with the best one receiving a score of 5, the second 3, the third 2, and 1 for the remaining results. We showed each user 6 scenes of different styles, and, in total, 25 participants were involved, yielding 150 rankings per evaluation aspect. Note that, for view consistency, we show rendered videos, while for style and content performance, we only display rendered views.} 
\label{tab:user_study}
\footnotesize
\setlength{\tabcolsep}{2pt} 
\begin{tabular}{ccccccc}
\toprule
 & StylizedGS & ABC-GS  & SGSST   & CLIPGaussian & StyleGaussian & Ours  \\ \midrule
Style        & 13.25\%    & 13.08\% & \underline{25.04\%} & 8.55\%       & 7.86\%        & \textbf{32.22\%} \\
Content      & 8.29\%     & \underline{19.04\%} & 10.26\% & 12.91\%      & 11.54\%       & \textbf{37.61\%} \\
Consistency     & 11.02\%    & \underline{19.83\%} & 11.03\% & 11.03\%      & 9.32\%        & \textbf{37.61\%} \\
\bottomrule
\end{tabular}
\end{table}

\subsubsection{Quantitative Results.}
We conduct quantitative comparisons in three aspects: stylization quality, content preservation, and view consistency. Specifically, we use ArtFID~\cite{artfid} to evaluate the stylization quality as in previous work. To evaluate content preservation, considering the substantial domain gap between stylized and real images, we adopt a cross-domain structural loss~\cite{splicing} to measure the structural similarity between the stylized and real scene. For multi-view consistency, we utilize MEt3R~\cite{met3r} to compute consistency error between two views at both short range (one-view gap) and long range (seven-view gap). We select five forward-facing scenes from LLFF~\cite{llff} and five 360-degree scenes from T\&T~\cite{tnt} and Mip-NeRF 360~\cite{mipnerf}, and stylize each scene using six different style images, producing 60 test cases per method. As evidenced in Table~\ref{tab:quant_eval}, our method outperforms baselines across all metrics. 

Since style transfer is inherently subjective, we also conducted a user study to assess human preferences. Our user study contains three evaluation aspects: style quality, content preservation, and multi-view consistency. For style quality and content preservation, we present the user with a rendering view before stylization, along with the corresponding stylized views produced by the five baselines and our approach, in random order. For view consistency, we show users rendered videos. A total of 25 participants have been involved, with 450 result rankings collected. The final preference ratio is summarized in Table~\ref{tab:user_study}. Again, our method leads the user study by a large margin, validating its high perceptual quality.

\begin{figure}[t]
\captionsetup[subfigure]{labelformat=empty, labelsep=none}
  \centering
  \begin{subfigure}{0.49\textwidth} 
    \centering
    \includegraphics[width=\textwidth]{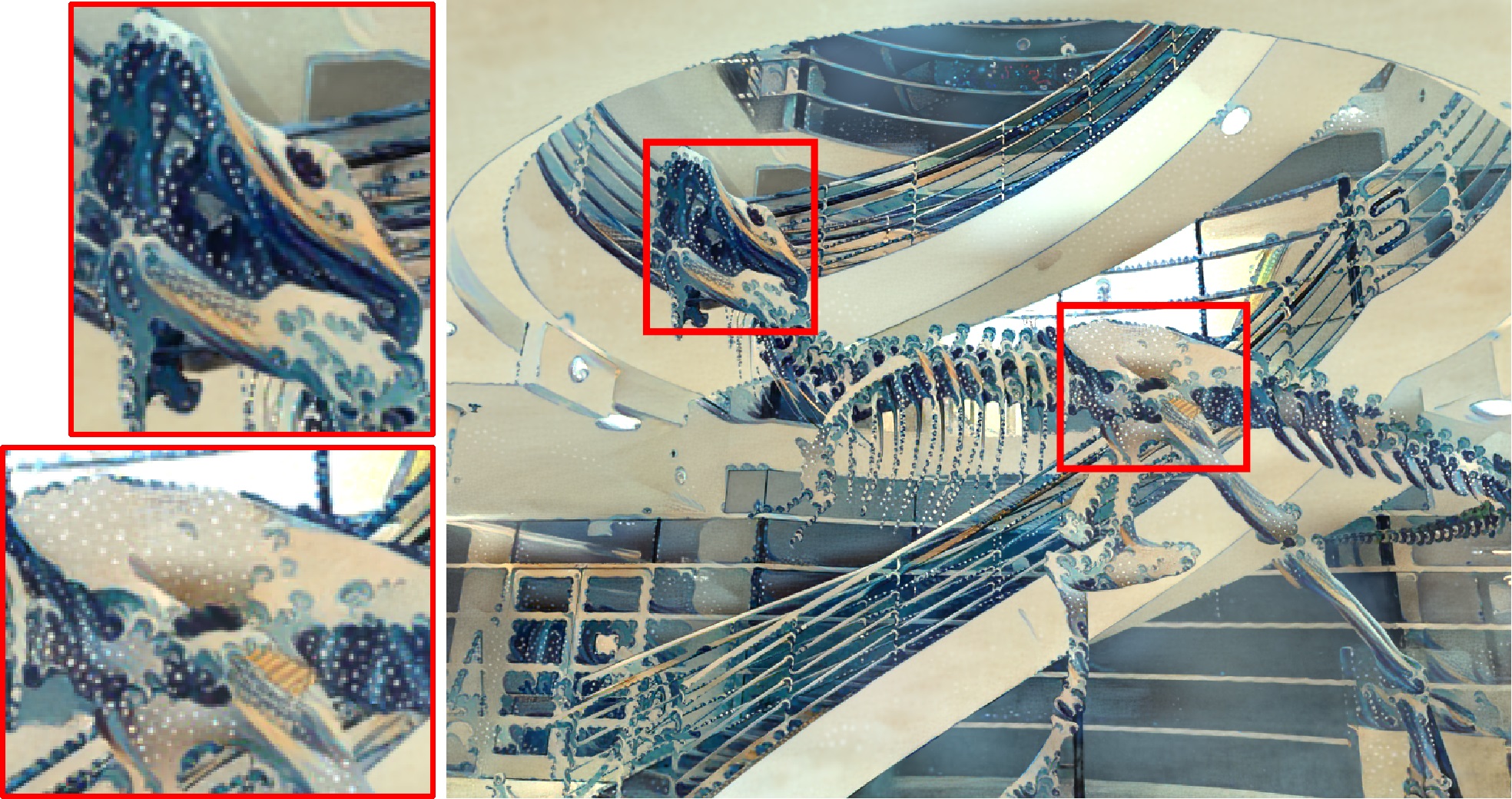}
    \captionsetup{oneside, margin={20mm,0mm}}
    \caption{\textbf{CCFT Loss}} 
  \end{subfigure}
  \hspace{-5pt}
  \begin{subfigure}{0.49\textwidth} 
    \centering
    \includegraphics[width=\textwidth]{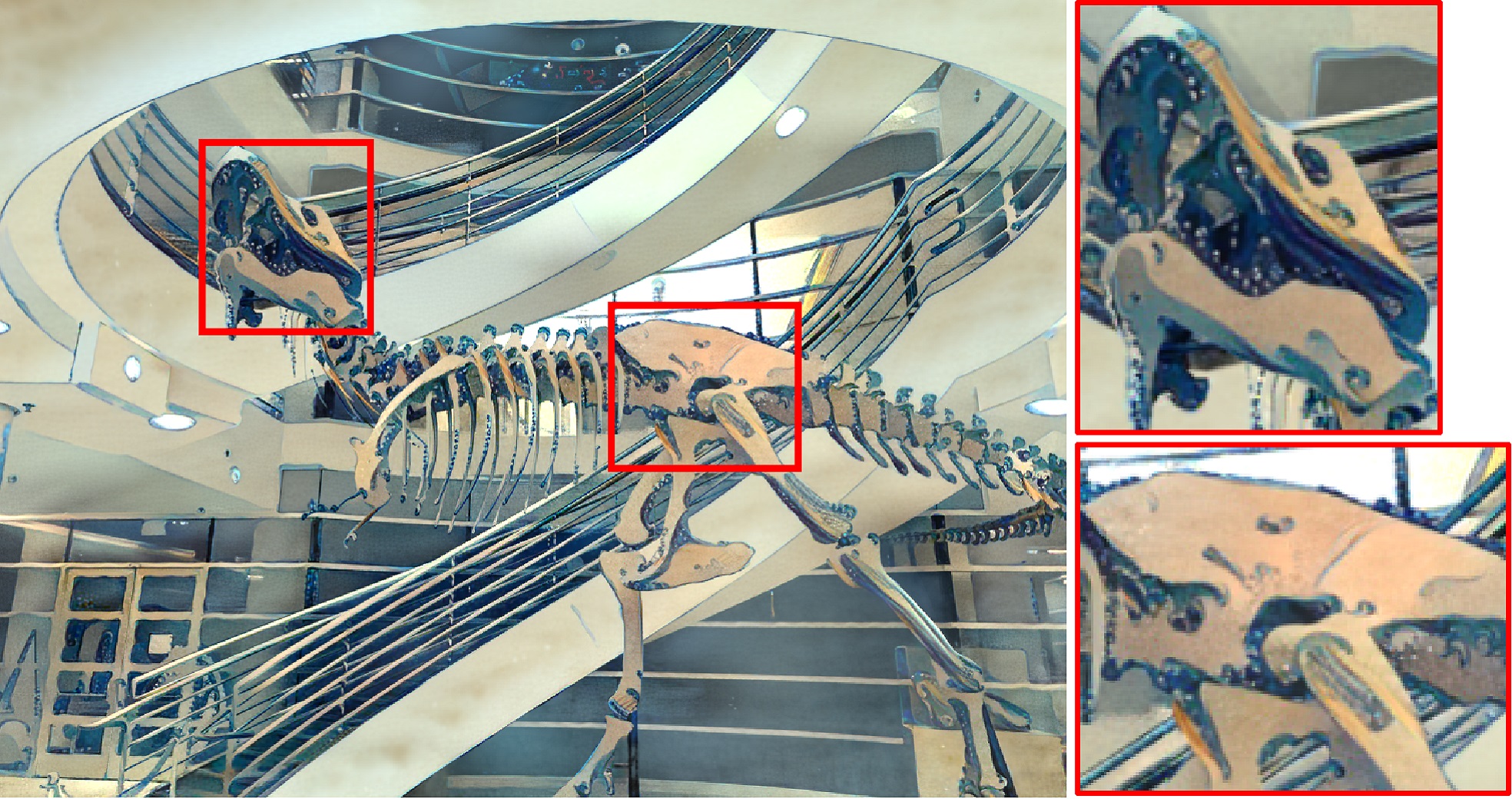}
    \captionsetup{oneside, margin={0mm,15mm}}
    \caption{\textbf{NNFM Loss}}
  \end{subfigure}
  \hspace{-5pt}
  \begin{subfigure}{0.49\textwidth} 
    \centering
    \includegraphics[width=\textwidth]{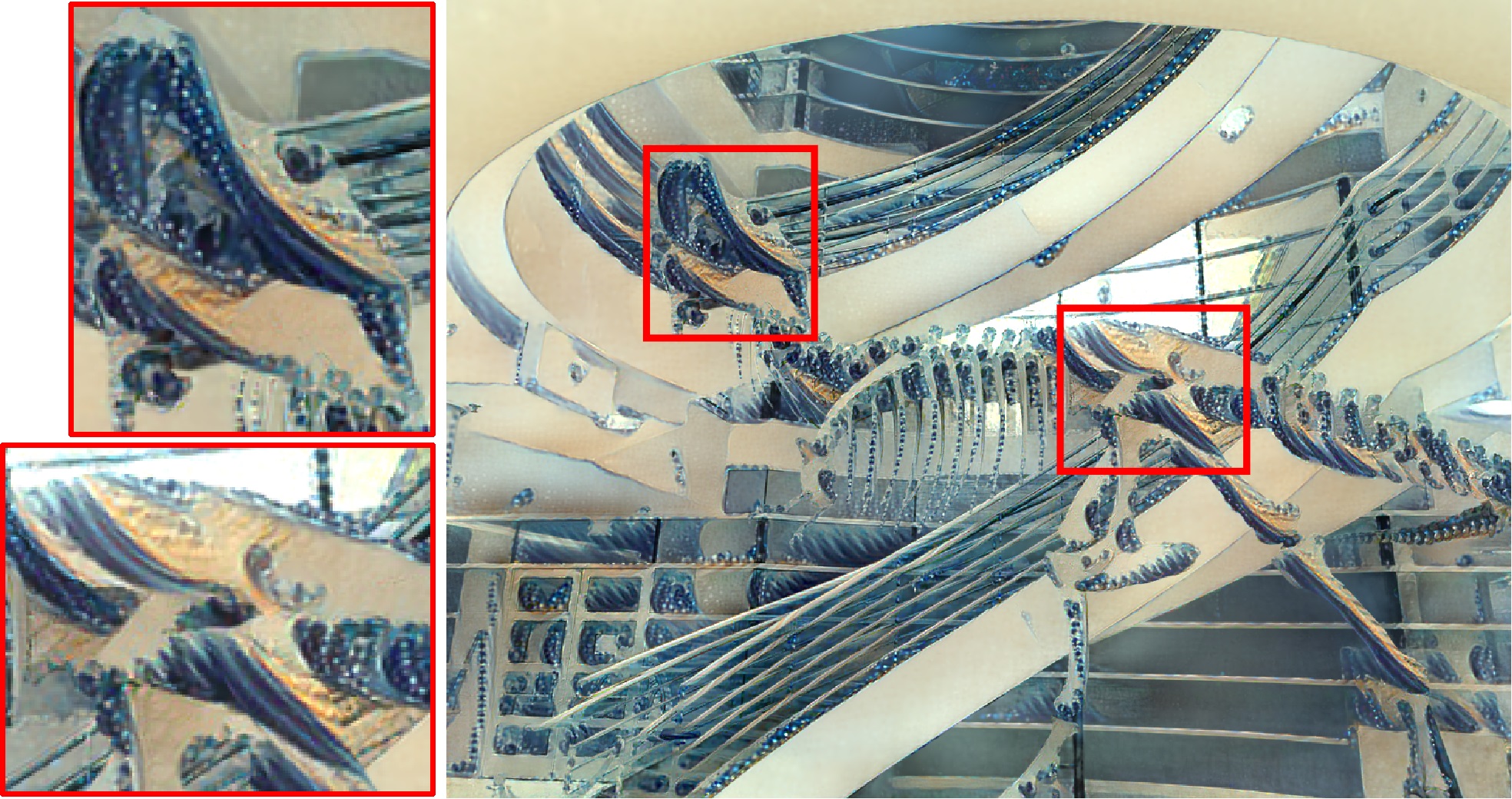}
    \captionsetup{oneside, margin={20mm,0mm}}
    \caption{\textbf{FAST Loss}} 
  \end{subfigure}
  \hspace{-5pt}
  \begin{subfigure}{0.49\textwidth} 
    \centering
    \includegraphics[width=\textwidth]{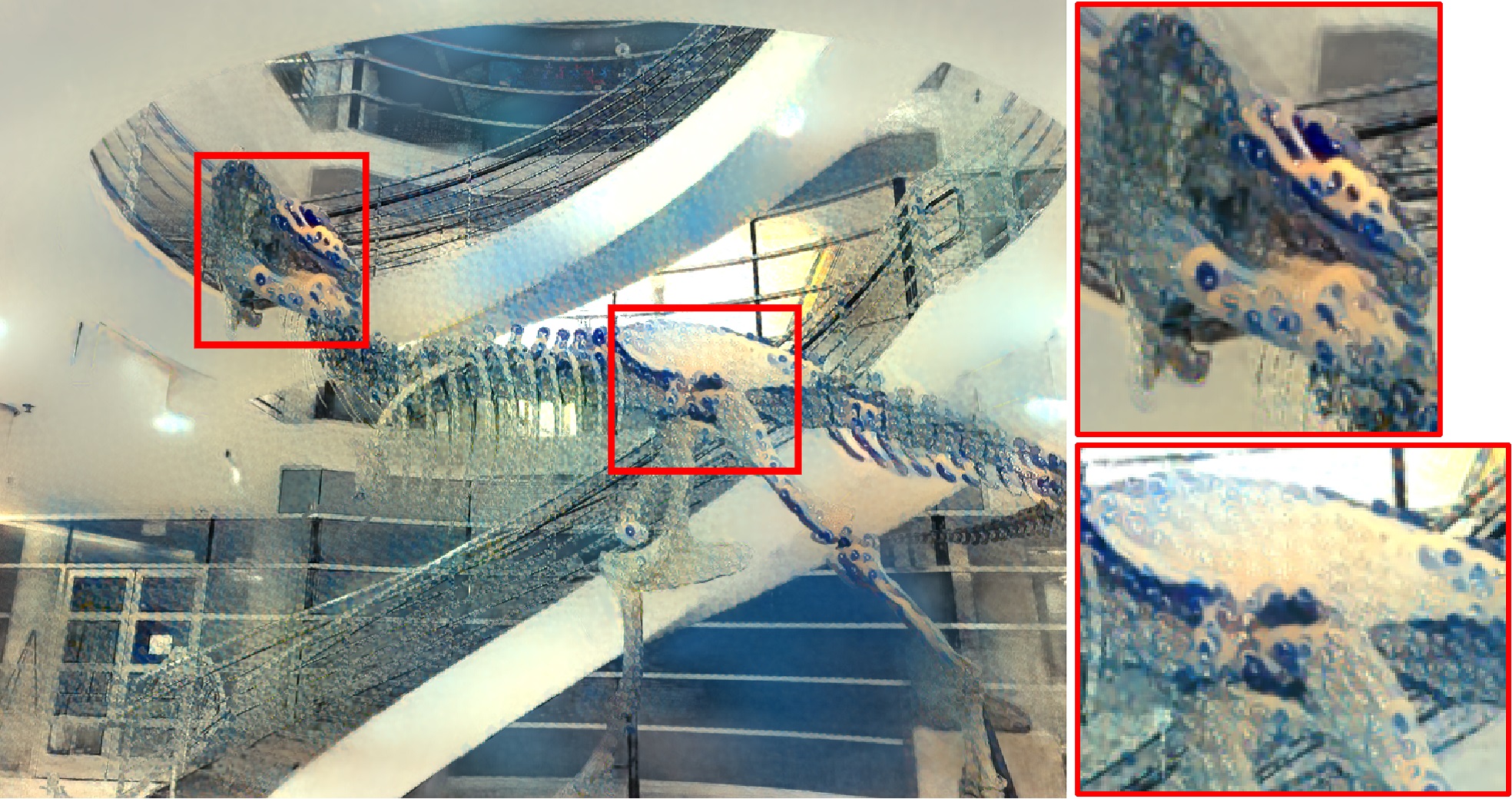}
    \captionsetup{oneside, margin={0mm,15mm}}
    \caption{\textbf{GRAM Loss}} 
  \end{subfigure}
  \hspace{-5pt}
  \caption{\textbf{Ablation study on loss functions.}}
  \label{fig:ccft_ablation}
\end{figure}

\begin{figure}[htbp]
\captionsetup[subfigure]{labelformat=empty, labelsep=none}
  \centering
  \begin{subfigure}{0.49\textwidth} 
    \centering
    \includegraphics[width=\textwidth]{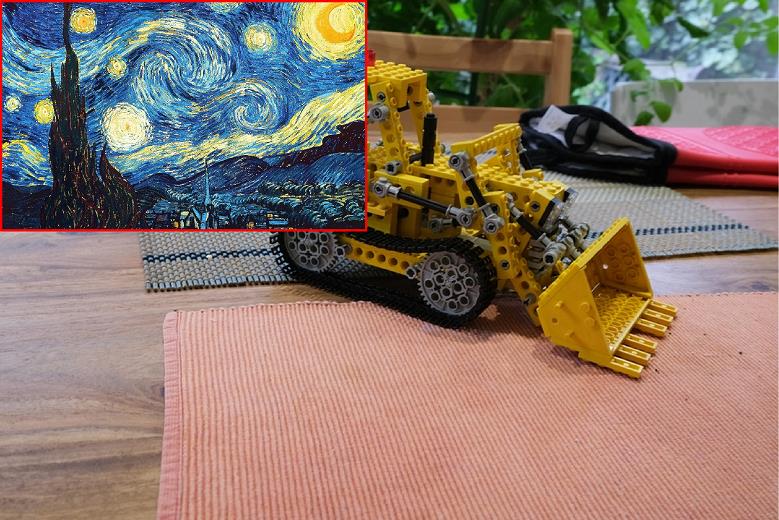}
    \caption{\textbf{Style $\&$ Content}} 
    \label{fig:sub1}
  \end{subfigure}
  \hspace{-5pt}
  \begin{subfigure}{0.49\textwidth} 
    \centering
    \includegraphics[width=\textwidth]{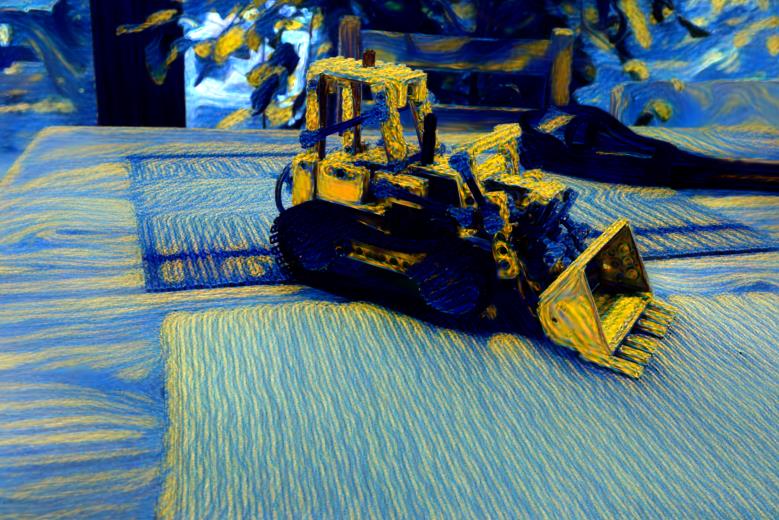}
    \caption{$\tau=0.01$}
    \label{fig:sub2}
  \end{subfigure}
  \hspace{-5pt}
  \begin{subfigure}{0.49\textwidth} 
    \centering
    \includegraphics[width=\textwidth]{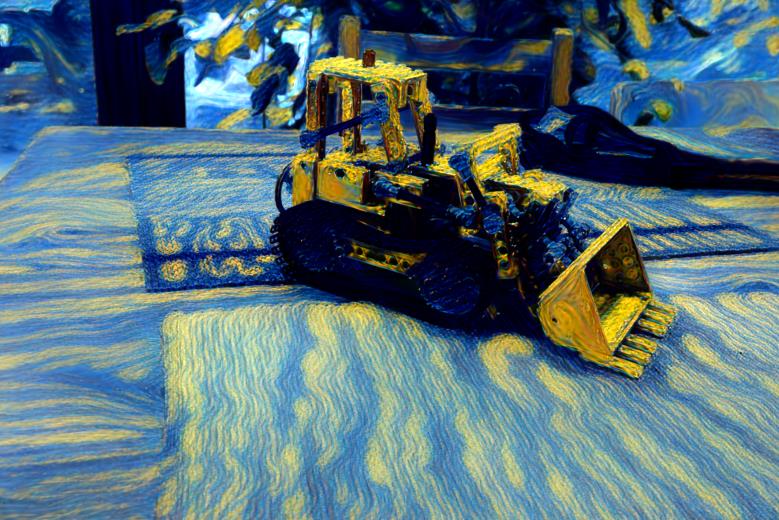}
    \caption{$\tau=0.1$} 
    \label{fig:sub1}
  \end{subfigure}
  \hspace{-5pt}
  \begin{subfigure}{0.49\textwidth} 
    \centering
    \includegraphics[width=\textwidth]{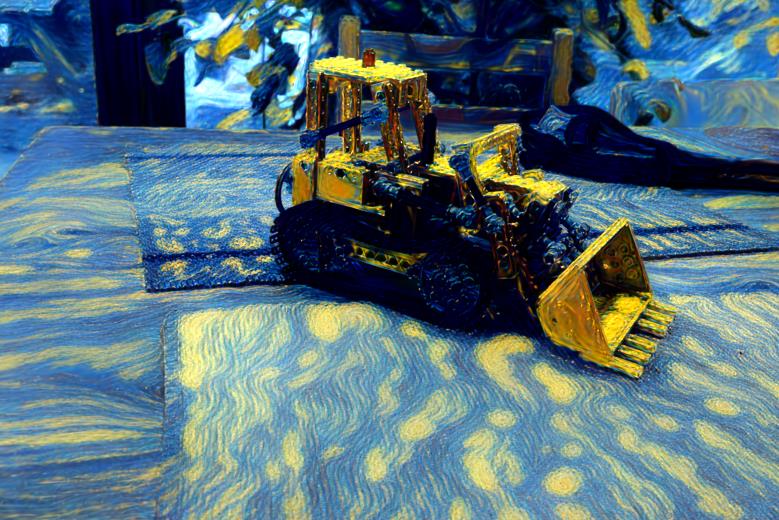}
    \caption{$\tau=1$} 
    \label{fig:sub1}
  \end{subfigure}
  \hspace{-5pt}
  \caption{\textbf{Ablation study for capacity-controlled feature transport.}}
  \label{fig:capacity_ablation}
\end{figure}

\subsection{Ablation Study}

\subsubsection{Loss function.}
We first compare our CCFT Loss to the NNFM Loss~\cite{arf}, FAST Loss~\cite{abc-gs} and GRAM Loss~\cite{nst}. Specifically, we replace the style loss with each alternative loss and use the style image and scene from the fifth column in Fig.~\ref{fig:qualitative}. As shown in Fig.~\ref{fig:ccft_ablation}, our CCFT loss successfully captures the sea-wave and water-splash patterns in the style image, faithfully preserving fine stylistic details. Although the FAST loss can also capture relevant style patterns, it produces less detailed results and causes artifacts. In contrast, the NNFM Loss only matches monotonous background patterns of the style image, while the GRAM Loss fails to transfer specific local style patterns and results more artifacts.

\subsubsection{Capacity control.}
To demonstrate the capacity control capability of our CCFT Loss, Fig.~\ref{fig:capacity_ablation} presents the stylization results with different capacity constraint strengths under the same style and scene. When the capacity constraint parameter $\tau$ is set to 0.01, the constraint is weak, and the transport plan tends to simply minimize the matching cost, assigning a large number of content features to a few lowest-cost style features, resulting in highly repetitive striped textures across large areas of the scene. When the parameter $\tau$ is increased to 0.1, the capacity constraint on lowest-cost style features is enhanced, which encourages content features to be distributed across a broader set of style features, leading to the emergence of swirl-like textures in the scene. When the parameter $\tau$ is further increased to 1, content features are assigned to style features in a more balanced manner, thereby introducing more detailed circular textures.

\subsubsection{Cross-view matching guidance.}
\begin{figure}[t]
\captionsetup[subfigure]{labelformat=empty, labelsep=none}
  \centering
  \begin{subfigure}{0.49\textwidth} 
    \centering
    \includegraphics[width=\textwidth]{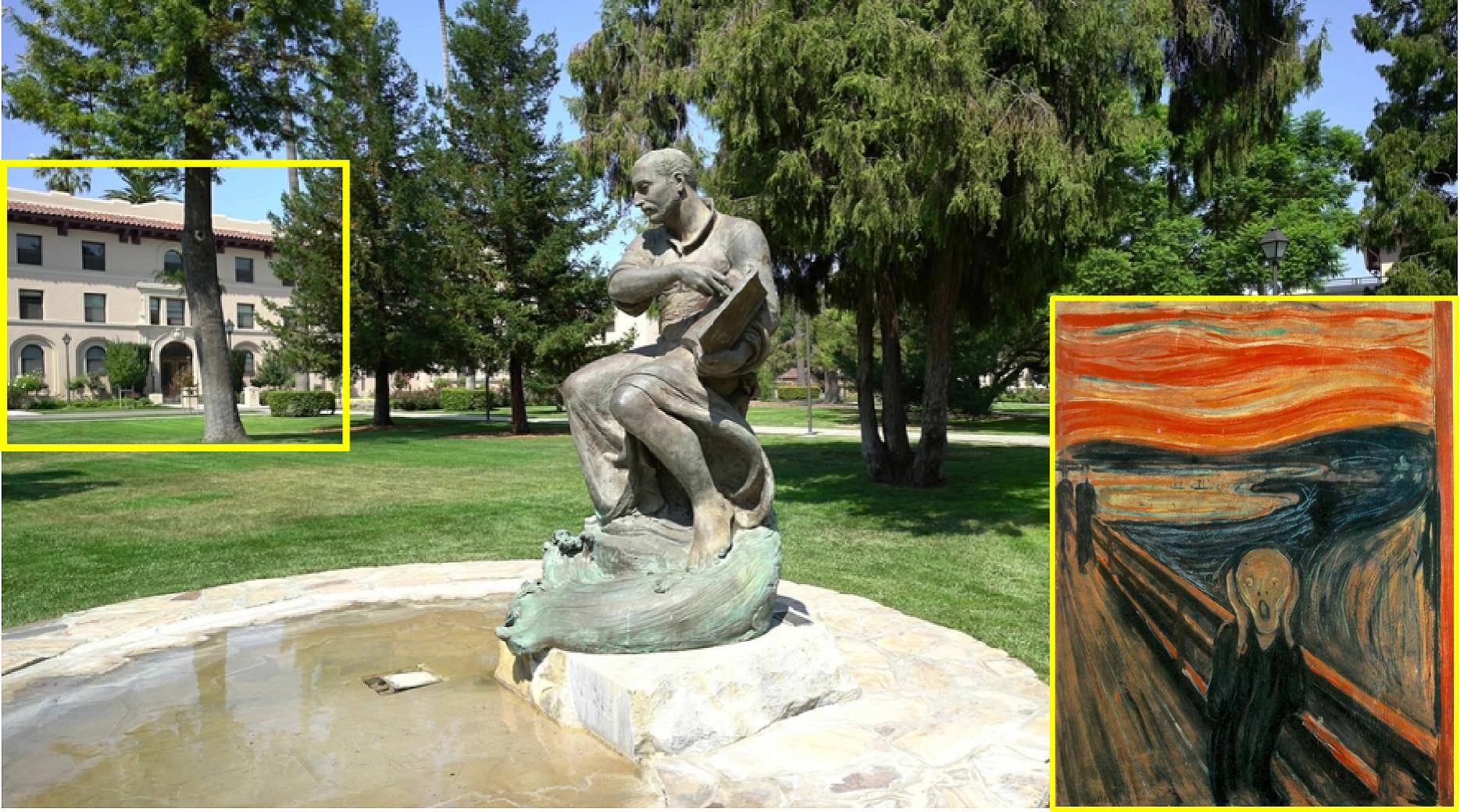}
    \caption{\textbf{Style $\&$ Content}} 
    \label{fig:sub1}
  \end{subfigure}
  \hspace{-5pt}
  \begin{subfigure}{0.49\textwidth} 
    \centering
    \includegraphics[width=\textwidth]{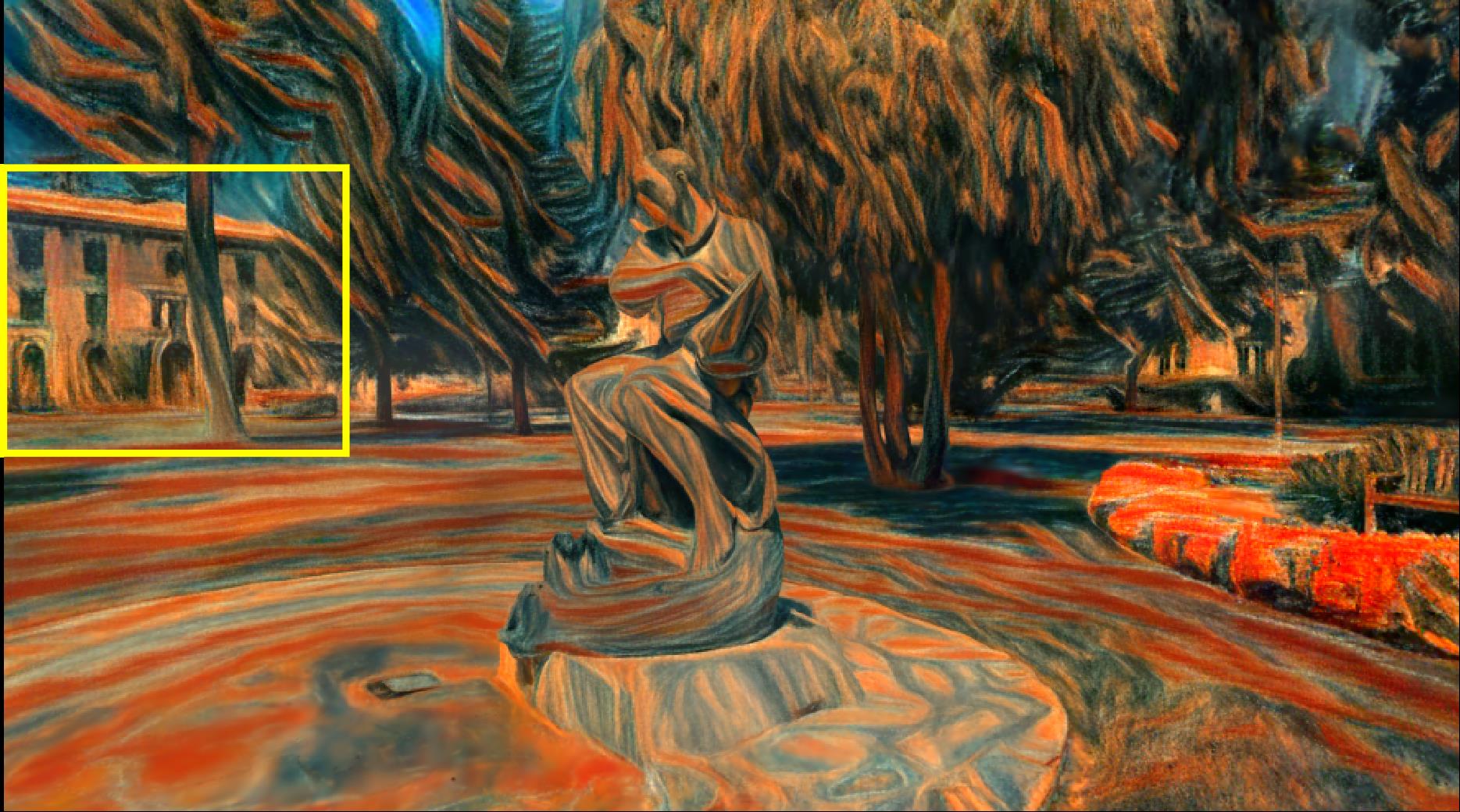}
    \caption{$\lambda_{guide}=0$}
    \label{fig:sub2}
  \end{subfigure}
  \hspace{-5pt}
  \begin{subfigure}{0.49\textwidth} 
    \centering
    \includegraphics[width=\textwidth]{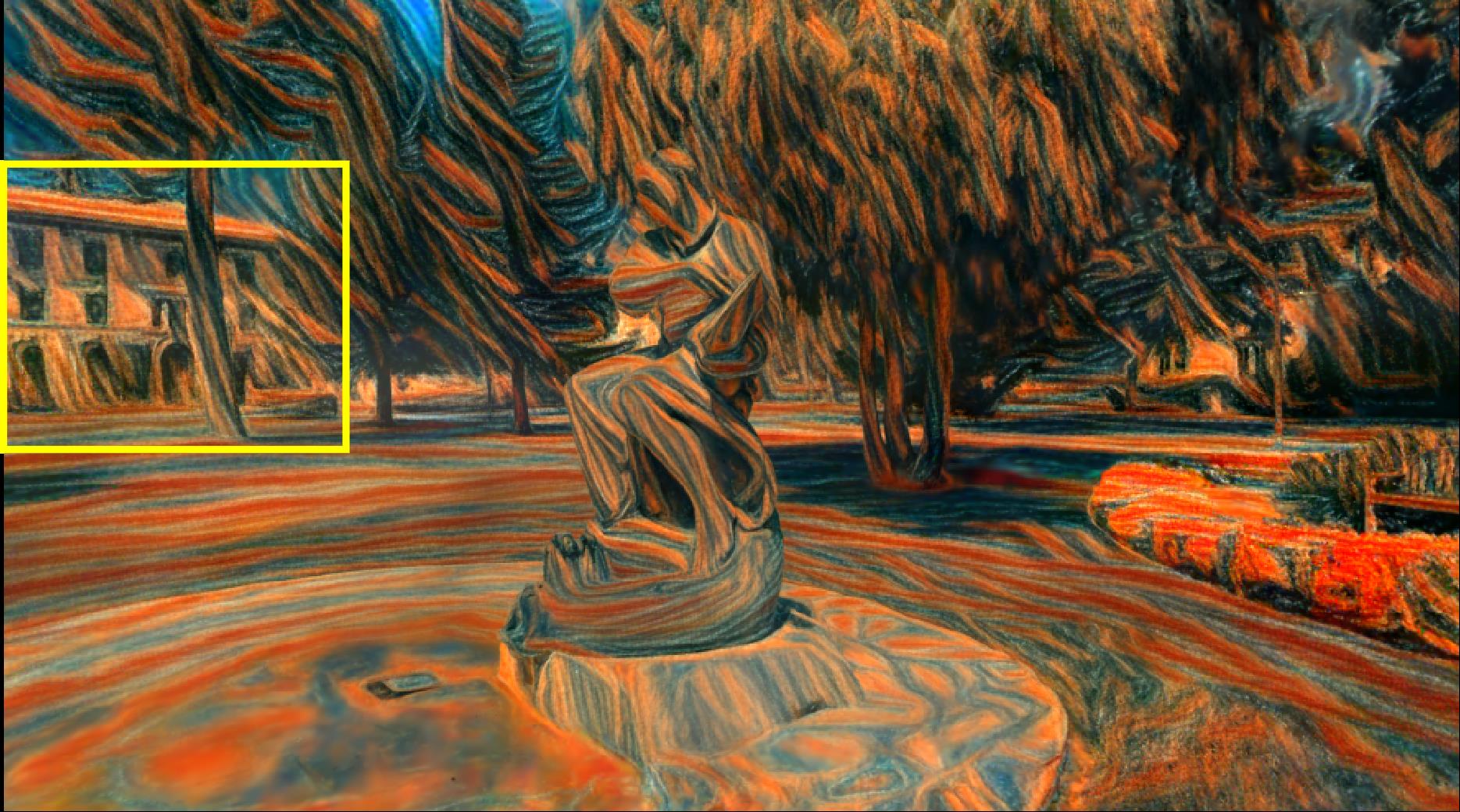}
    \caption{$\lambda_{guide}=1$} 
    \label{fig:sub1}
  \end{subfigure}
  \hspace{-5pt}
  \begin{subfigure}{0.49\textwidth} 
    \centering
    \includegraphics[width=\textwidth]{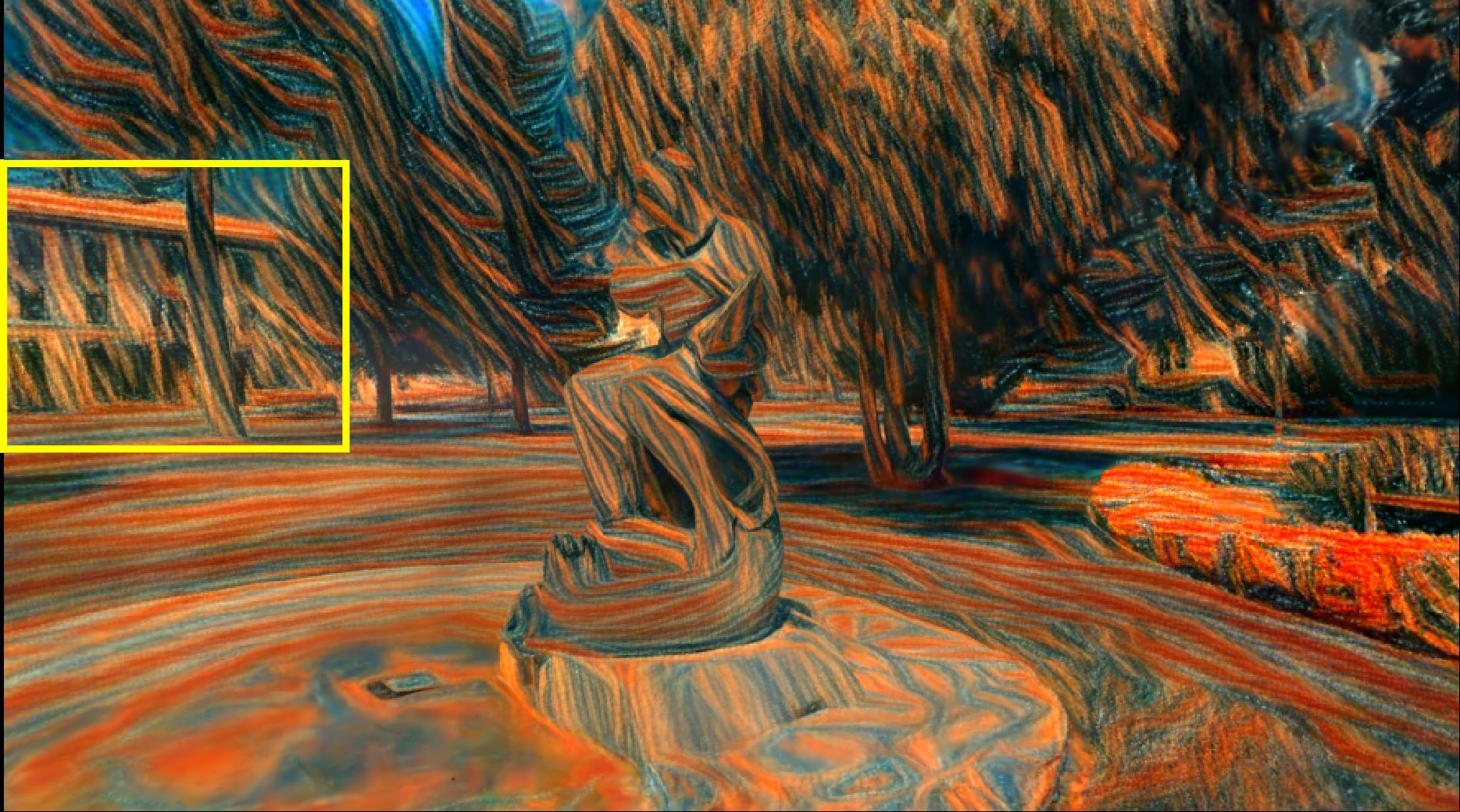}
    \caption{$\lambda_{guide}=10$} 
    \label{fig:sub1}
  \end{subfigure}
  \hspace{-5pt}
  \caption{\textbf{Ablation on different strengths of cross-view matching guidance.}}
  \label{fig:guide_ablation}
\end{figure}

We evaluate the effectiveness of our cross‑view guidance in improving multi‑view consistency through additional quantitative experiments. Table~\ref{tab:guide_ablation} reports the qualitative results of the cross‑view guidance under different parameter settings. The results show that as the strength of guidance increases, both short‑range and long‑range consistency exhibit improvements. This indicates that cross‑view guidance can effectively mitigate inconsistent textures arising from viewpoint changes during multi‑view training, aligning with our expectations. However, we also observe that as the parameter is gradually increased, the fidelity of the scene structure declines slightly. To explain this phenomenon, we further analyze the stylization results in Fig.~\ref{fig:guide_ablation}. Essentially, this trade‑off in content stems from the “2D–3D gap” between 2D views and 3D scenes: cross‑view guidance tends to cover part of the content details in the scene with consistent 2D style patterns, making the projected textures of the 3D scene appear more consistent across viewpoints, which ``damages'' some original content boundaries. Therefore, in practice, we set the guidance weight to a reasonable value ($\lambda_{guide}=1$) during stylization. 

\subsubsection{Enhanced reconstruction.}
To evaluate the contribution of our enhanced reconstruction, we use the original 3DGS with our capacity-controlled stylization. As shown in Tab.~\ref{tab:quant_eval}, this ablated version achieves quite competitive results, especially in multi-view consistency, compared to the baselines. We further present a more detailed ablation study of each constraint in the supplementary.

\begin{table}[t]
\centering
\caption{Ablation study for cross-view matching guidance.}
\label{tab:guide_ablation}
\footnotesize
\setlength{\tabcolsep}{2.5pt} 
\begin{tabular}{lccccc}
\toprule


$\lambda_{guide}$ & ArtFID$\downarrow$ & Structure Loss$\downarrow$ & Short MEt3R$\downarrow$ & Long MEt3R$\downarrow$ \\

\midrule
0 & 22.809 & \textbf{0.0312} & 0.1204 & 0.2807 \\
1 & \textbf{22.801} & \underline{0.0318} & \underline{0.1196} & \underline{0.2795} \\
10 & \underline{22.805} & 0.0323 & \textbf{0.1191} & \textbf{0.2786} \\

\bottomrule
\end{tabular}
\end{table}

\section{Conclusions}
\label{sec:conclusions}
In this paper, we present a capacity-controlled framework for multi-view stylization of 3D Gaussian Splatting. Addressing multi-view inconsistency in existing 3DGS stylization methods, we propose a Capacity-Controlled Feature Transport (CCFT) loss that prevents many-to-one matching while enabling tunable control over style diversity. Furthermore, we introduce cross-view matching guidance to constrain the same 3D content to match consistent style patterns. We also develop an enhanced scene reconstruction with geometric regularizations to establish a robust foundation for high-quality stylization. 

While our method produces superior stylization results, several limitations remain. First, the cross-view matching guidance introduces a trade-off between multi-view consistency and content preservation. Second, our method requires per-scene optimization. Future work will explore feed-forward approaches to enable real-time stylization while preserving the controllability and consistency benefits of our framework.


\section*{Acknowledgements}
This work was supported in part by ICFCRT (W2441020), Guangdong Basic and Applied Basic Research Foundation (2023B1515120026), SZU Teaching Reform Research Project (JG2026011), and Guangdong Provincial Key Laboratory of Visual Media and Multidimensional Intelligence.

%
%
\bibliographystyle{splncs04}
\bibliography{main}

\newpage
\appendix
\clearpage
\setcounter{page}{1}

\section{Supplementary Material}
This supplementary material contains additional details or results about our implementation, runtime \& memory, ablation study, user study, and our semi-balanced optimal transport. A supplementary video with voiceover narration that compares our method to the baselines is available at \textcolor{magenta}{\url{https://youtu.be/V3welUPVOao}}.

\subsection{Additional Implementation Details }
\label{sec:view_iter}
For enhanced scene reconstruction, we use the default settings of the original 3DGS~\cite{3dgs}, so the training time is nearly the same as that of the original 3DGS. For scene stylization, as we run the optimization view-by-view sequentially, we determine the number of training iterations based on the number of views in a scene: for forward-facing scenes, we perform 50 iterations per view with a minimum constraint of 2,000 total iterations; for 360° scenes with 200 or fewer views, we apply 20 iterations per view, whereas for 360° scenes exceeding 200 views, we apply 15 iterations per view.

\subsection{Runtime and Memory}
\label{sec:time}
We compare the stylization efficiency of other methods in terms of runtime and peak GPU memory consumption. On average, StylizedGS requires 1.45 minutes and 1.03 GB of GPU memory, while ABC-GS takes 5.38 minutes and 9.63 GB. Our method completes stylization in 8.35 minutes with a peak memory usage of 8.37 GB. The additional runtime of our method primarily stems from the iterative Sinkhorn optimization and the cross-view guidance mechanism used to enforce fine-grained, view-consistent stylization. In contrast, the sparse-support computation strategy adopted in CCFT reduces memory usage compared with ABC-GS, resulting in a lower peak GPU memory footprint. Overall, although our method introduces moderate computational overhead, it achieves significantly higher stylization quality and cross-view consistency, yielding a favorable trade-off between efficiency and performance.

\subsection{Additional Ablation Study}
\label{sec:regularization}

\subsubsection{Geometric regularizations.}
\begin{figure}[t]
  \centering
  \begin{subfigure}{0.49\textwidth} 
    \centering
    \includegraphics[width=\textwidth]{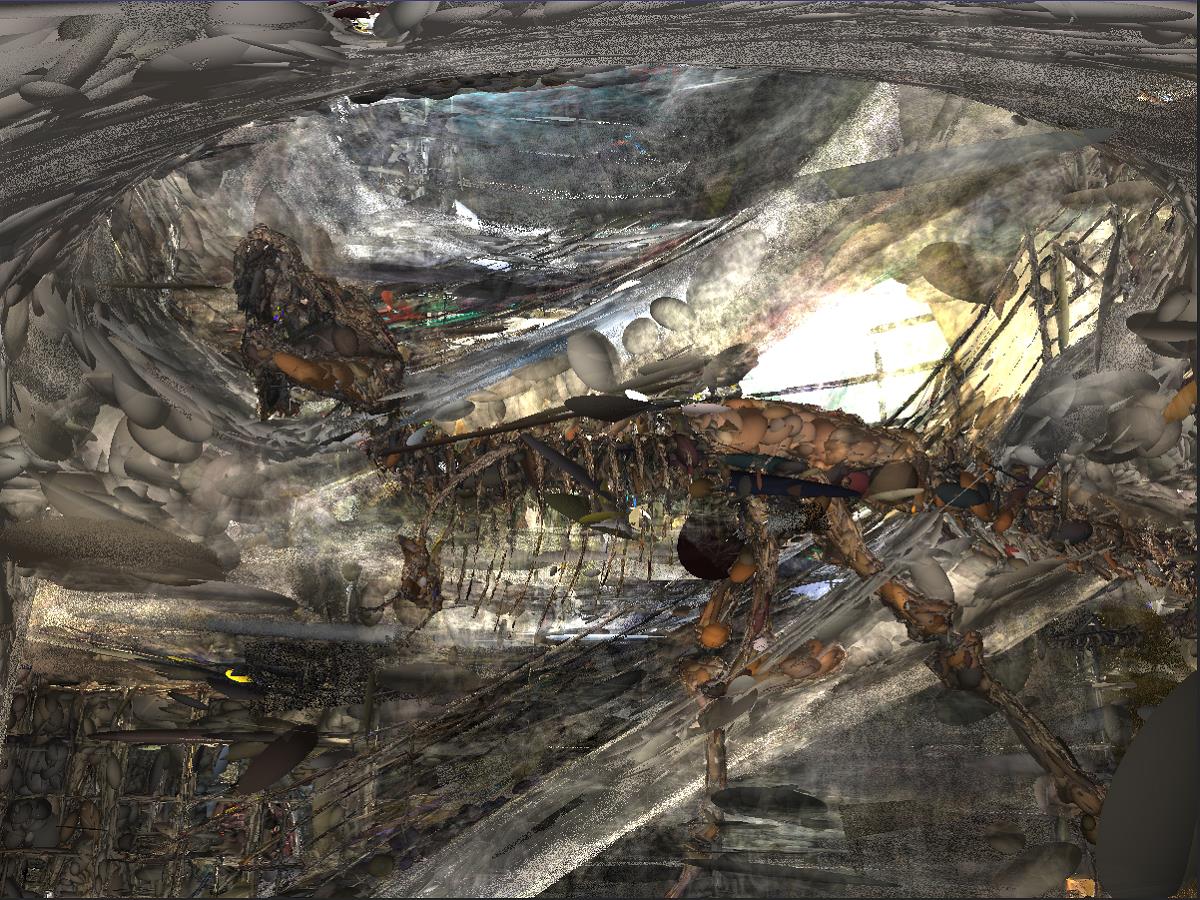}
    \caption{$\mathcal{L}_{\mathit{photo}}$} 
    \label{fig:geometric_ablation_a}
  \end{subfigure}
  \hspace{-5pt}
  \begin{subfigure}{0.49\textwidth} 
    \centering
    \includegraphics[width=\textwidth]{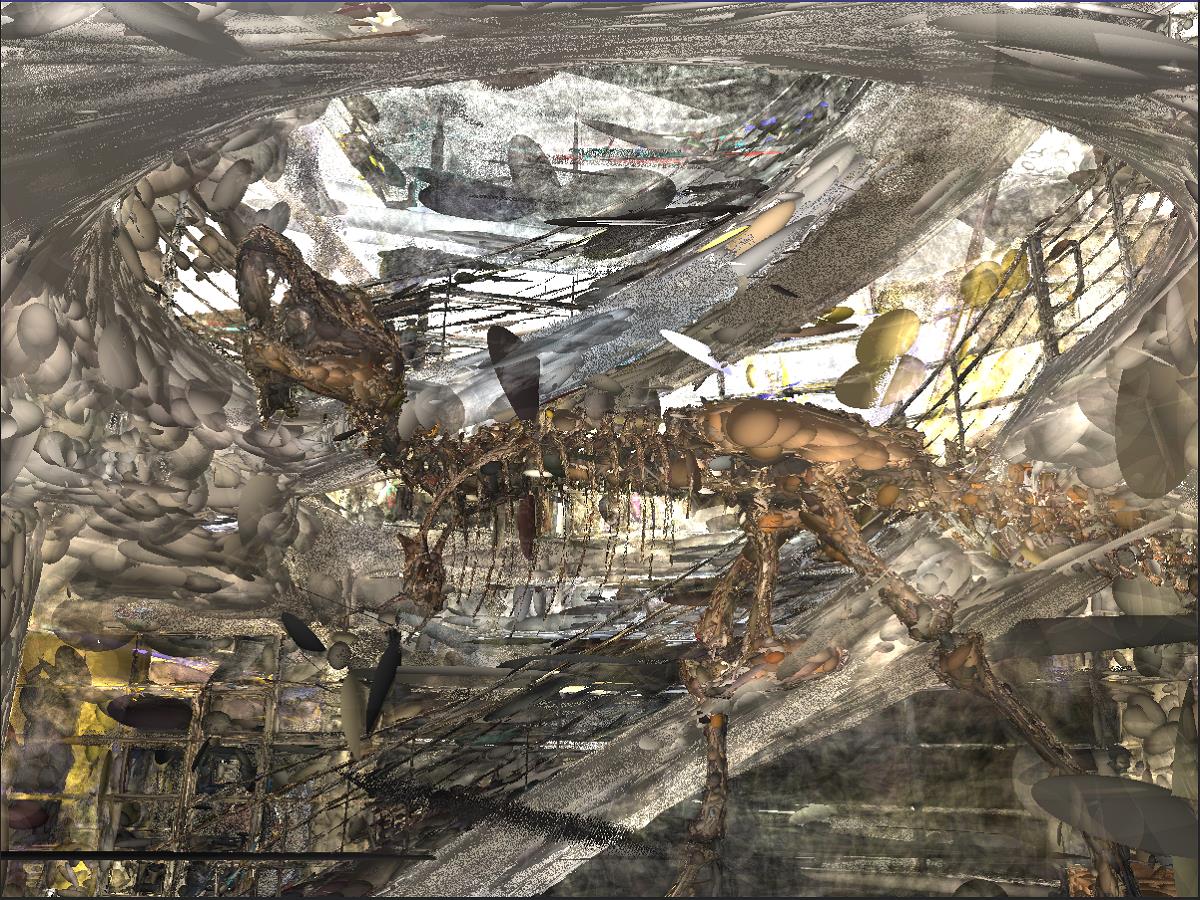}
    \caption{$\mathcal{L}_{\mathit{photo}}+\mathcal{L}_{\mathit{depth}}$}
    \label{fig:sub2}
  \end{subfigure}
  \hspace{-5pt}
  
  \vspace{2pt}
  
  \begin{subfigure}{0.49\textwidth} 
    \centering
    \includegraphics[width=\textwidth]{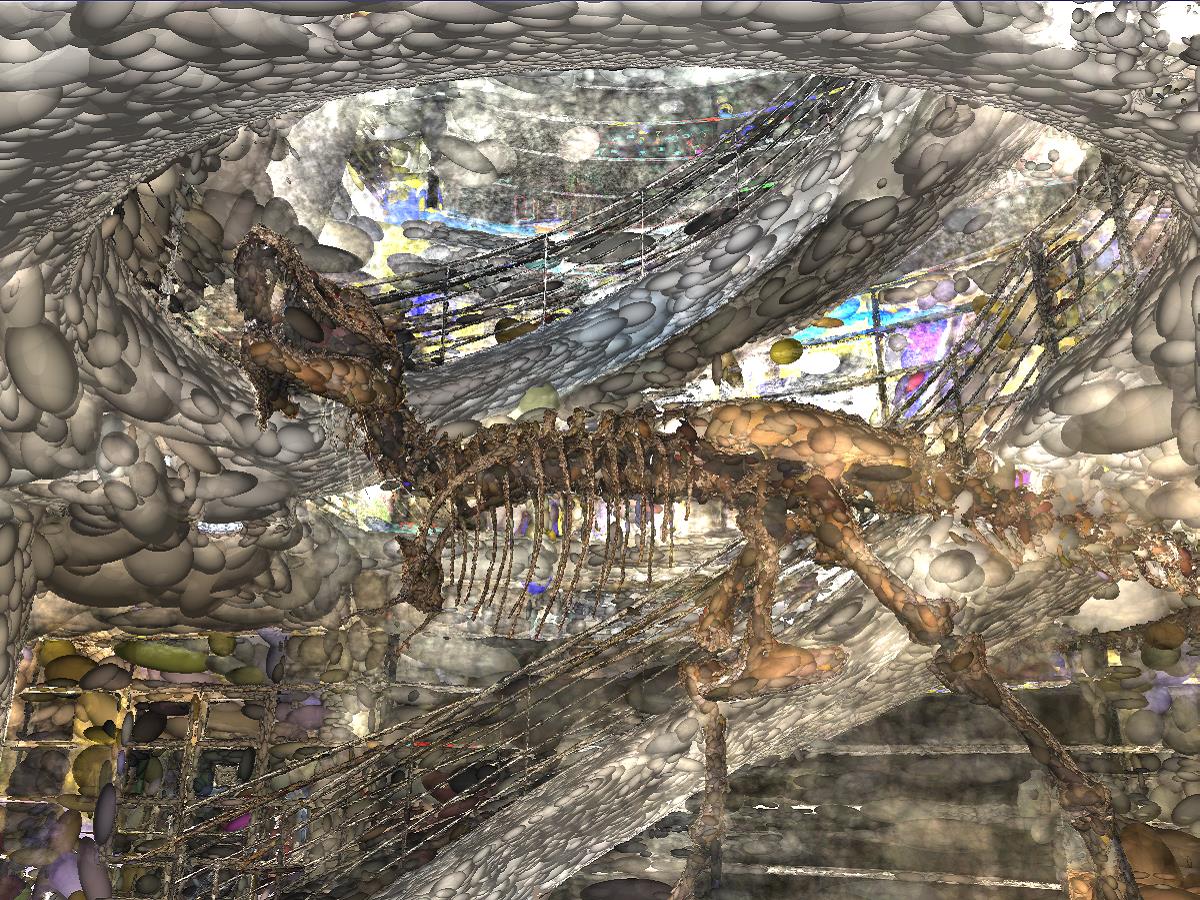}
    \caption{$\mathcal{L}_{\mathit{photo}}+\mathcal{L}_{\mathit{depth}}+\mathcal{L}_{\mathit{shape}}$} 
    \label{fig:sub1}
  \end{subfigure}
  \hspace{-5pt}
  \begin{subfigure}{0.49\textwidth} 
    \centering
    \includegraphics[width=\textwidth]{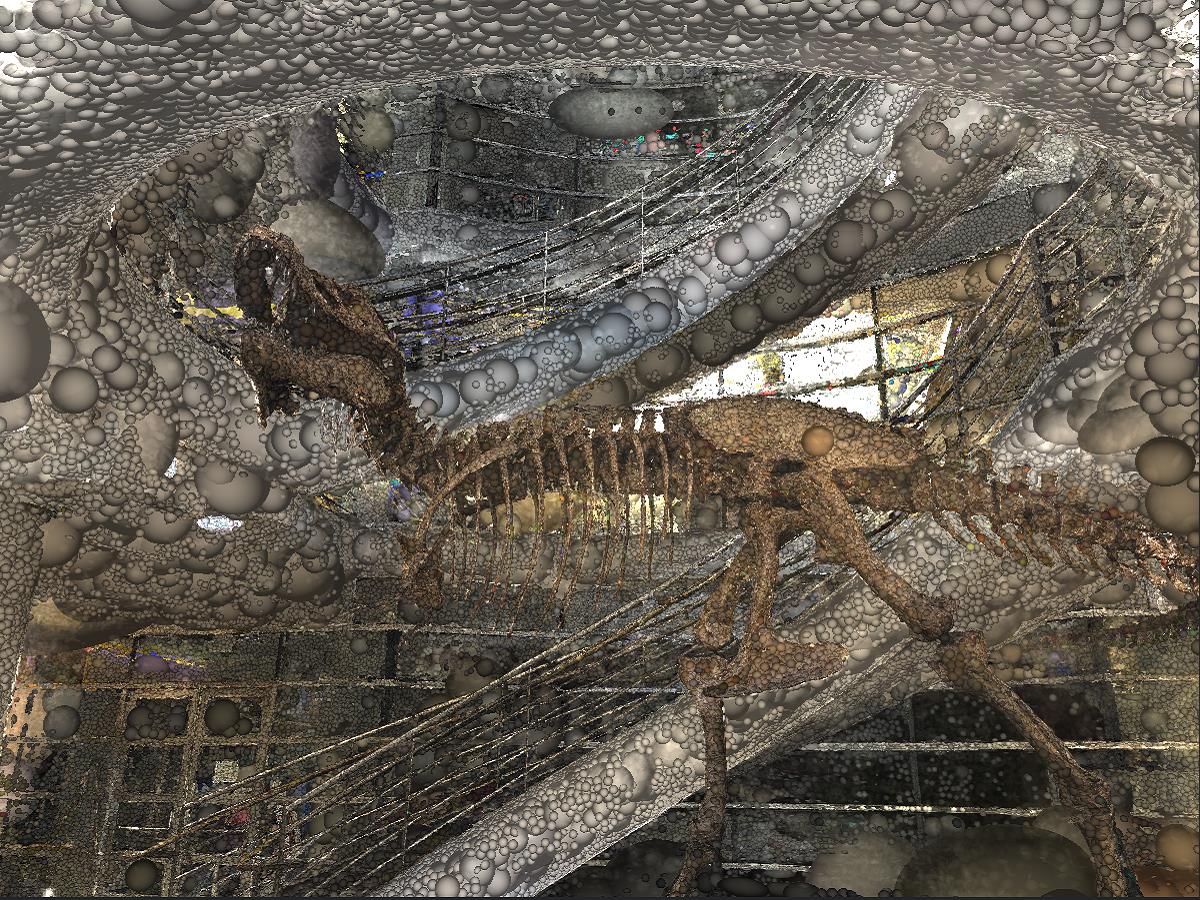}
    \caption{$\mathcal{L}_{\mathit{photo}}+\mathcal{L}_{\mathit{depth}}+\mathcal{L}_{\mathit{shape}}+\mathcal{L}_{\mathit{size}}$} 
    \label{fig:sub1}
  \end{subfigure}
  \hspace{-5pt}
  \caption{\textbf{Ablation study of geometric regularizations.}}
  \label{fig:geometric_ablation}
\end{figure}

We present the ablation results of the geometric regularization terms in Fig.~\ref{fig:geometric_ablation}, where each subfigure visualizes the Gaussian primitives in the reconstructed scene. As observed in Fig.~\ref{fig:geometric_ablation}\:$\mathbf{(a)}$, when reconstructing the scene using solely $\mathcal{L}_{\mathit{photo}}$, although 3DGS successfully fits the color of the real view, the overall geometric structure remains chaotic, and numerous Gaussian primitives exhibit disordered expansion and erroneous aggregation. Fig.~\ref{fig:geometric_ablation}\:$\mathbf{(b)}$ demonstrates that the incorporation of $\mathcal{L}_{\mathit{depth}}$ stabilizes the global spatial structure of the scene. The relative positions of the dinosaur skeleton, the ground plane, and the background become more physically plausible; however, some Gaussian primitives with irregular shapes persist. As shown in Fig.~\ref{fig:geometric_ablation}\:$\mathbf{(c)}$, further incorporating $\mathcal{L}_{\mathit{shape}}$ effectively suppresses abnormally elongated, needle-shaped, or highly anisotropic Gaussian primitives, leading to more uniform surface coverage and more regular structures, especially on walls and surrounding regions. Finally, the addition of $\mathcal{L}_{\mathit{size}}$ further refines the Gaussian primitives. The reconstruction results shown in Fig.~\ref{fig:geometric_ablation}\:$\mathbf{(d)}$ exhibit finer detail resolution and structural clarity, with the texture and contour of the scene becoming more prominent. In summary, these four sets of results indicate that $\mathcal{L}_{\mathit{photo}}$ provides the fundamental appearance-fitting capability, $\mathcal{L}_{\mathit{depth}}$ enhances global geometric consistency, $\mathcal{L}_{\mathit{shape}}$ further improves the structural rationality of the Gaussian primitives, and $\mathcal{L}_{\mathit{size}}$ effectively boosts detail recovery and local clarity.

\subsubsection{Enhanced reconstruction.}
\begin{figure}
  \centering
  \begin{subfigure}{0.49\textwidth} 
    \centering
    \includegraphics[width=\textwidth]{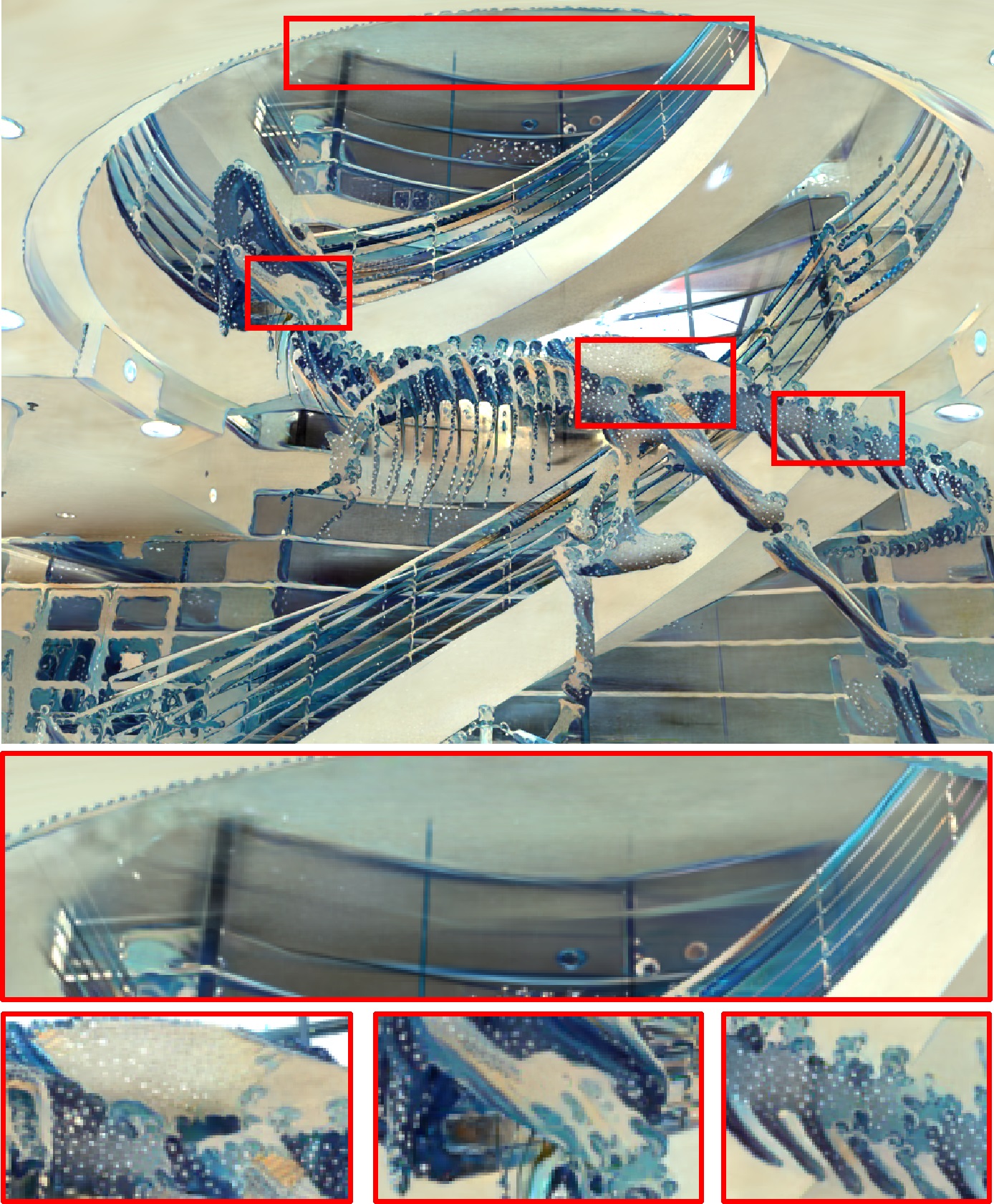}
    \caption{\textbf{Original Reconstruction}} 
    \label{fig:sub1}
  \end{subfigure}
  \hspace{-5pt}
  \begin{subfigure}{0.49\textwidth} 
    \centering
    \includegraphics[width=\textwidth]{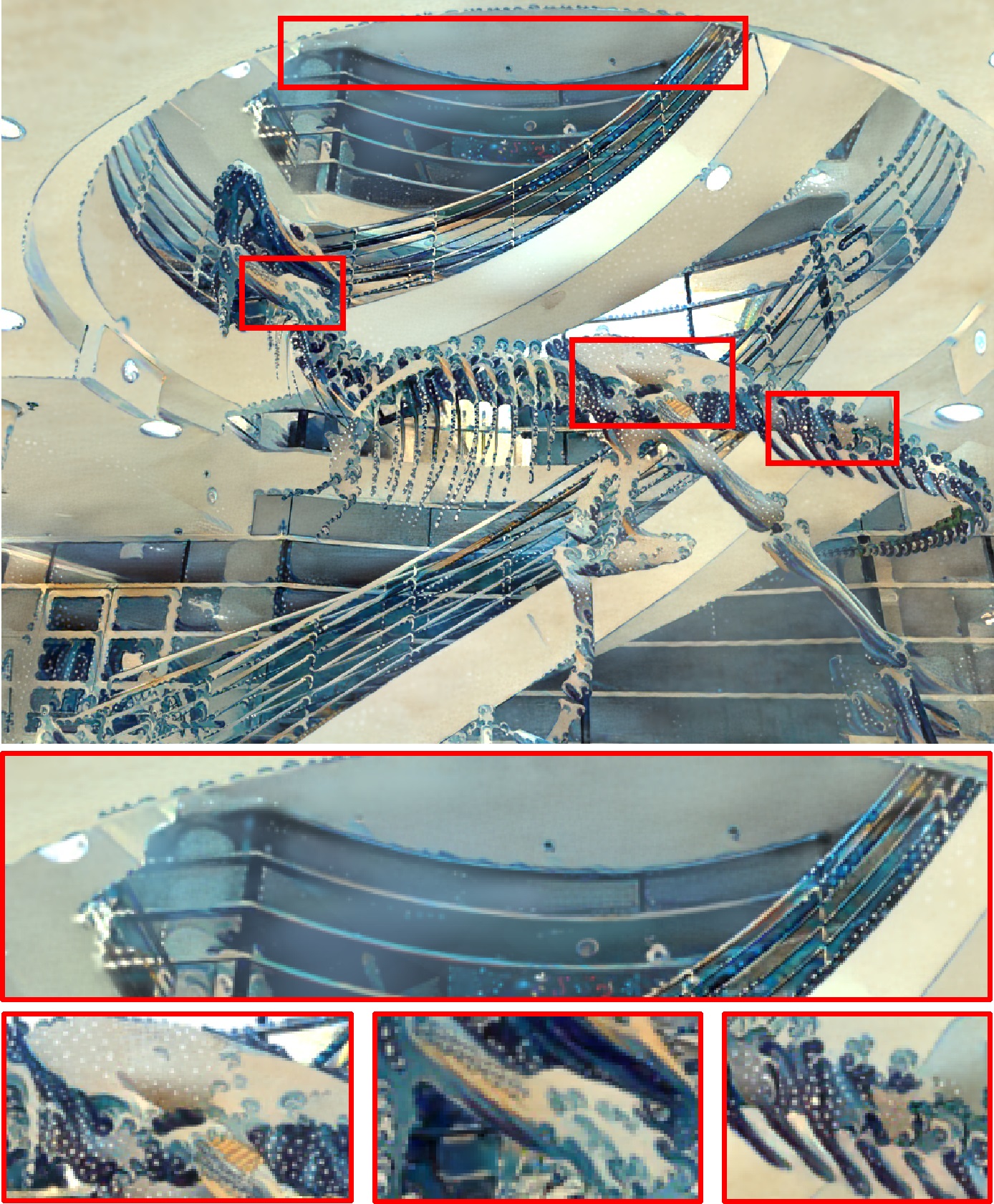}
    \caption{\textbf{Enhanced Reconstruction}}
    \label{fig:sub2}
  \end{subfigure}
  \hspace{-5pt}
  \caption{\textbf{Ablation study of enhanced reconstruction.}}
  \label{fig:enhanced_ablation_1}
\end{figure}

\begin{figure}
  \raggedright
  \begin{subfigure}{0.49\textwidth} 
    \centering
    \includegraphics[width=\textwidth]{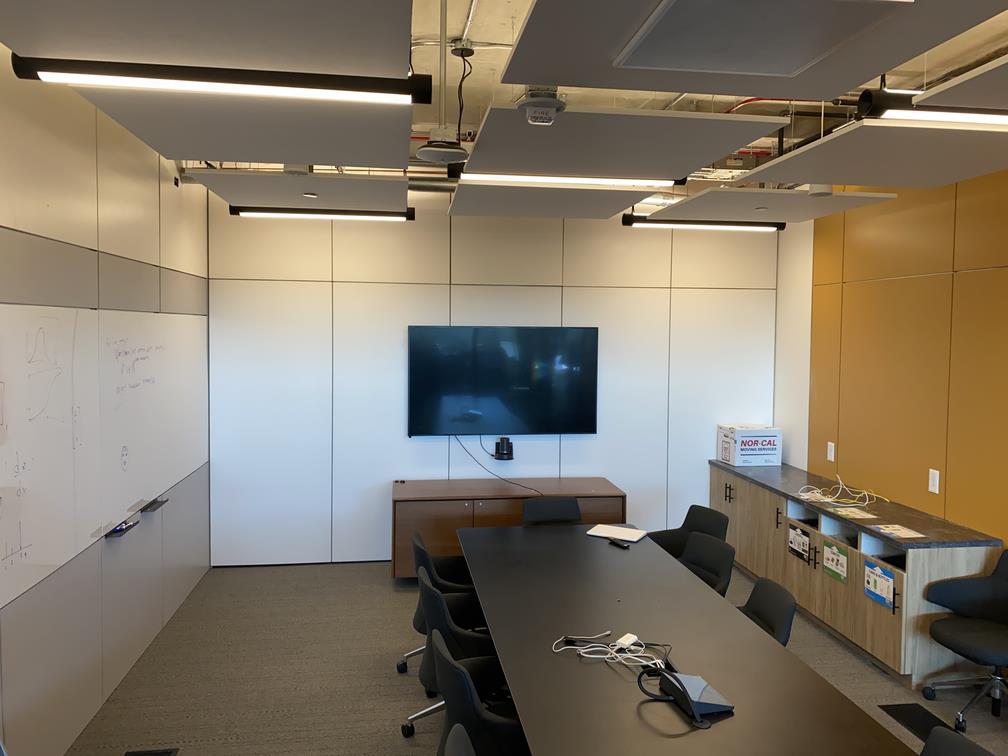}
    \caption*{\textbf{Content}} 
    \label{fig:sub1}
  \end{subfigure}
  \hspace{-5pt}
  \begin{subfigure}{0.207\textwidth} 
    \centering
    \includegraphics[width=\textwidth]{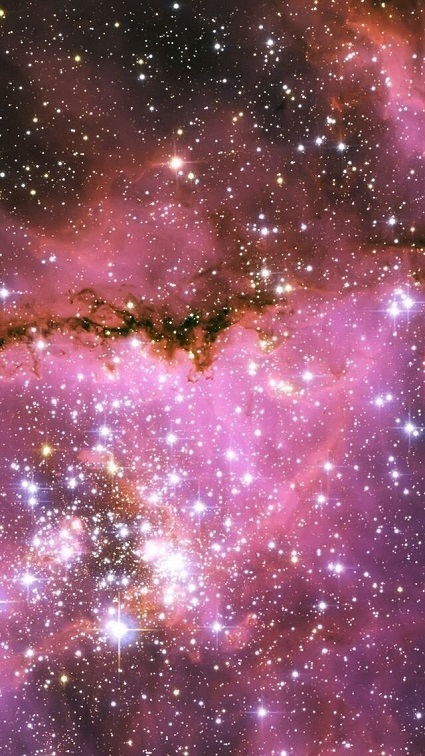}
    \caption*{\textbf{Style}}
    \label{fig:sub2}
  \end{subfigure}
  \hspace{-5pt}
  \begin{subfigure}{0.49\textwidth} 
    \centering
    \includegraphics[width=\textwidth]{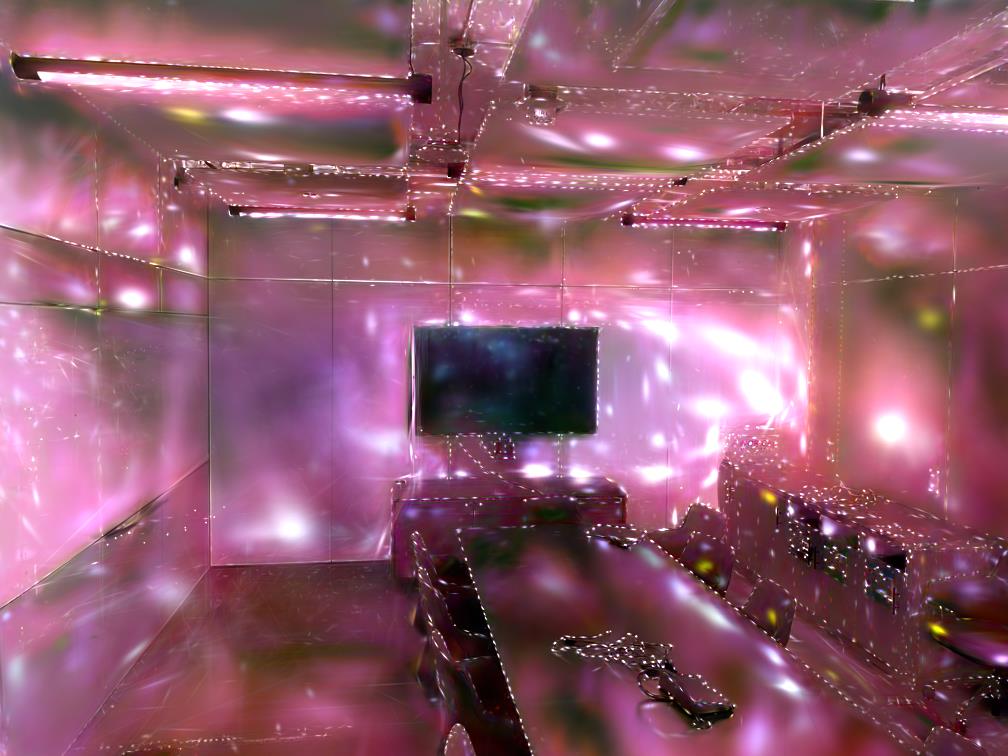}
    \caption{\textbf{Original Reconstruction}}
    \label{fig:sub1}
  \end{subfigure}
  \hspace{-5pt}
  \begin{subfigure}{0.49\textwidth} 
    \centering
    \includegraphics[width=\textwidth]{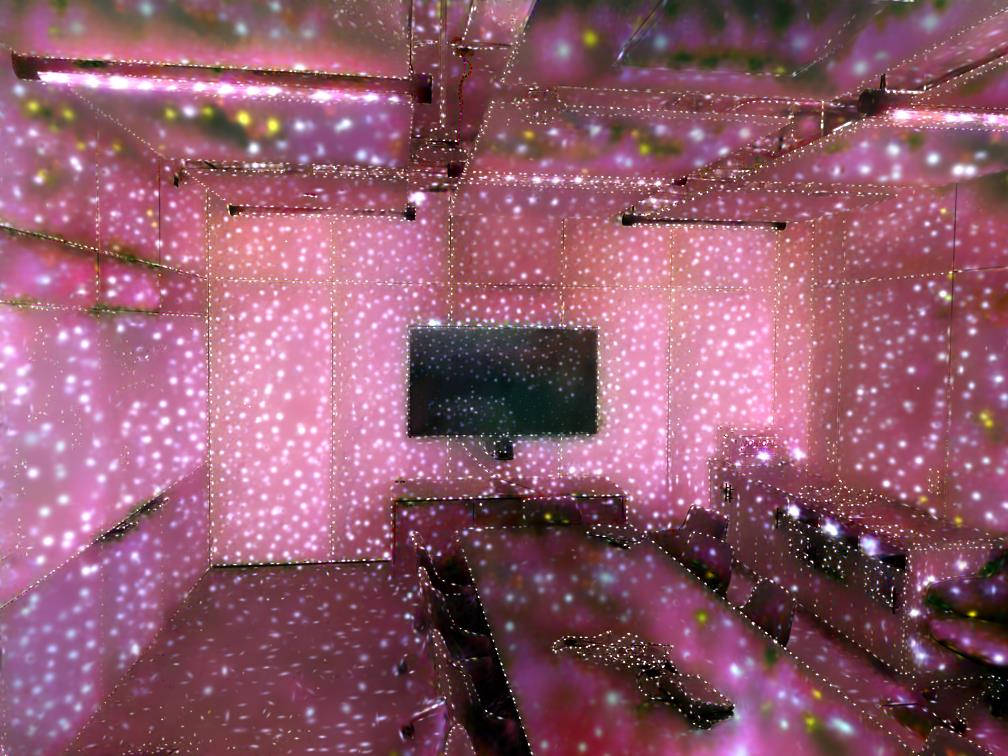}
    \caption{\textbf{Enhanced Reconstruction}}
    \label{fig:sub1}
  \end{subfigure}
  \hspace{-5pt}
  \caption{\textbf{Ablation study of enhanced reconstruction.}}
  \label{fig:enhanced_ablation_2}
\end{figure}

To validate the specific contribution of the proposed enhanced reconstruction to the stylization quality, we conducted an ablation study by comparing the stylization results based on enhanced reconstruction and original reconstruction. As observed in Fig.~\ref{fig:enhanced_ablation_1}\:$\mathbf{(a)}$, when relying solely on the original reconstruction method for stylization, the geometric structures and local texture details of the scene exhibit degradation and blurring. In ceiling and circular railings, the line edges appear severely smudged, resulting in indistinguishable geometry and structural discontinuities, while in dinosaur skeletons, stylized patterns suffer from excessive smoothness and loss of local details. In contrast, the results obtained with enhanced reconstruction exhibit consistently better visual quality in these regions. In Fig.~\ref{fig:enhanced_ablation_1}\:$\mathbf{(b)}$, our method maintains high coherence and sharpness in the railing structures and architectural edges; more importantly, the stylized texture distribution around the dinosaur skeleton regions conforms more faithfully to the underlying structure, and preserves local details more effectively. In Fig.~\ref{fig:enhanced_ablation_2}, the results of enhanced reconstruction exhibit a more significant improvement in style quality. The high-frequency granular textures of the style image are more faithfully preserved and distributed across walls, ceilings, and desk surfaces, while the original geometric silhouettes of the indoor scene remain sharp. This demonstrates that our method not only enhances geometric accuracy, but also provides a more stable supporting substrate for detailed style patterns.

\begin{table}[t]
\centering
\caption{Ablation study for training view order.}
\label{tab:view_abla}
\footnotesize
\setlength{\tabcolsep}{2.5pt} 
\begin{tabular}{lcccc}
\toprule


 & ArtFID$\downarrow$ & Structure Loss$\downarrow$ & Short MEt3R$\downarrow$ & Long MEt3R$\downarrow$ \\

\midrule
w/o guide & 22.809 & \textbf{0.0312} & 0.1204 & 0.2807 \\
random view & 22.806 & 0.0315 & 0.1202 & 0.2804 \\
ordered view & \textbf{22.801} & 0.0318 & \textbf{0.1196} & \textbf{0.2795} \\

\bottomrule
\end{tabular}
\end{table}

\subsubsection{View order and spatial relationship.}
Although our cross-view matching guidance provides an effective way to enforce view-consistent stylization, it assumes that adjacent views for training are captured in a spatially continuous manner with ordered overlap. This assumption naturally holds for the forward-facing and 360$^\circ$ datasets used in our experiments. To further examine its impact, we additionally conducted experiments using randomly ordered views. As shown in Tab.~\ref{tab:view_abla}, the guidance effect is significantly weakened when the view order is randomized, resulting in performance that is close to the variant without cross-view guidance. For custom datasets, particularly those with sparse or non-uniform camera trajectories, spatially continuous views can be obtained via \emph{camera interpolation} or by inserting \emph{virtual viewpoints}, thereby restoring the effectiveness of the proposed guidance mechanism. We leave this as a future direction.

\begin{figure}
\captionsetup[subfigure]{labelformat=empty, labelsep=none}
  \centering
  \begin{subfigure}{0.49\textwidth} 
    \centering
    \includegraphics[width=\textwidth]{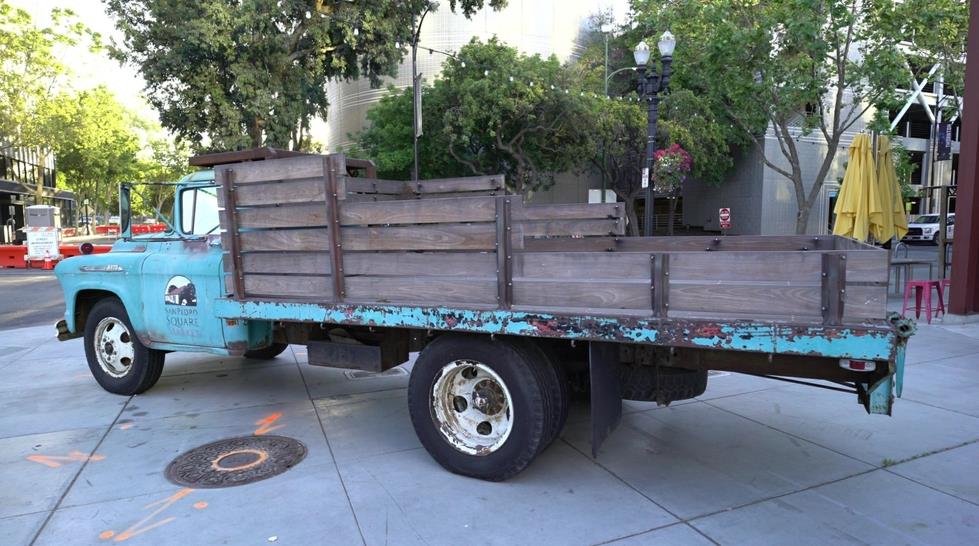}
    \caption{\textbf{viewpoint 1}} 
    \label{fig:sub1}
  \end{subfigure}
  \hspace{-5pt}
  \begin{subfigure}{0.49\textwidth} 
    \centering
    \includegraphics[width=\textwidth]{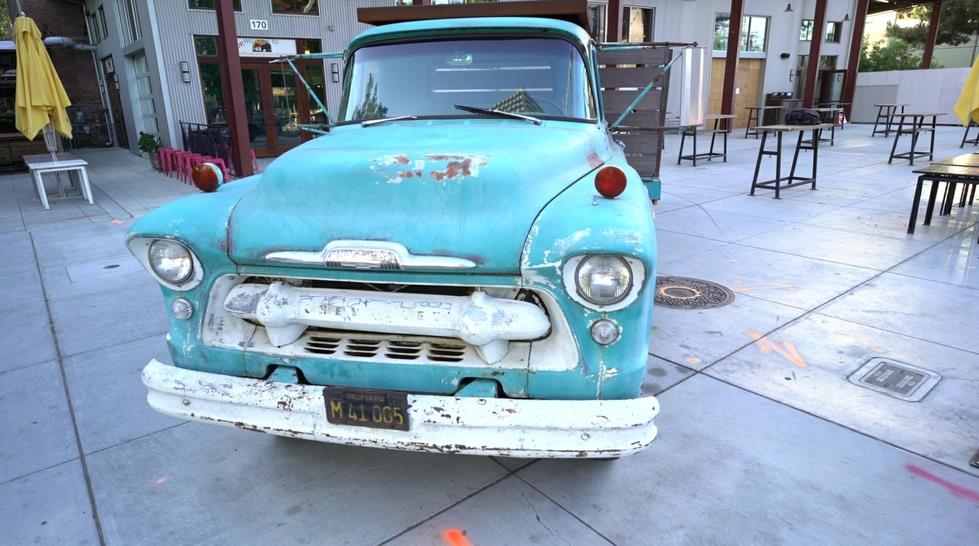}
    \caption{\textbf{viewpoint 2}}
    \label{fig:sub2}
  \end{subfigure}
  \hspace{-5pt}
  \begin{subfigure}{0.49\textwidth} 
    \centering
    \includegraphics[width=\textwidth]{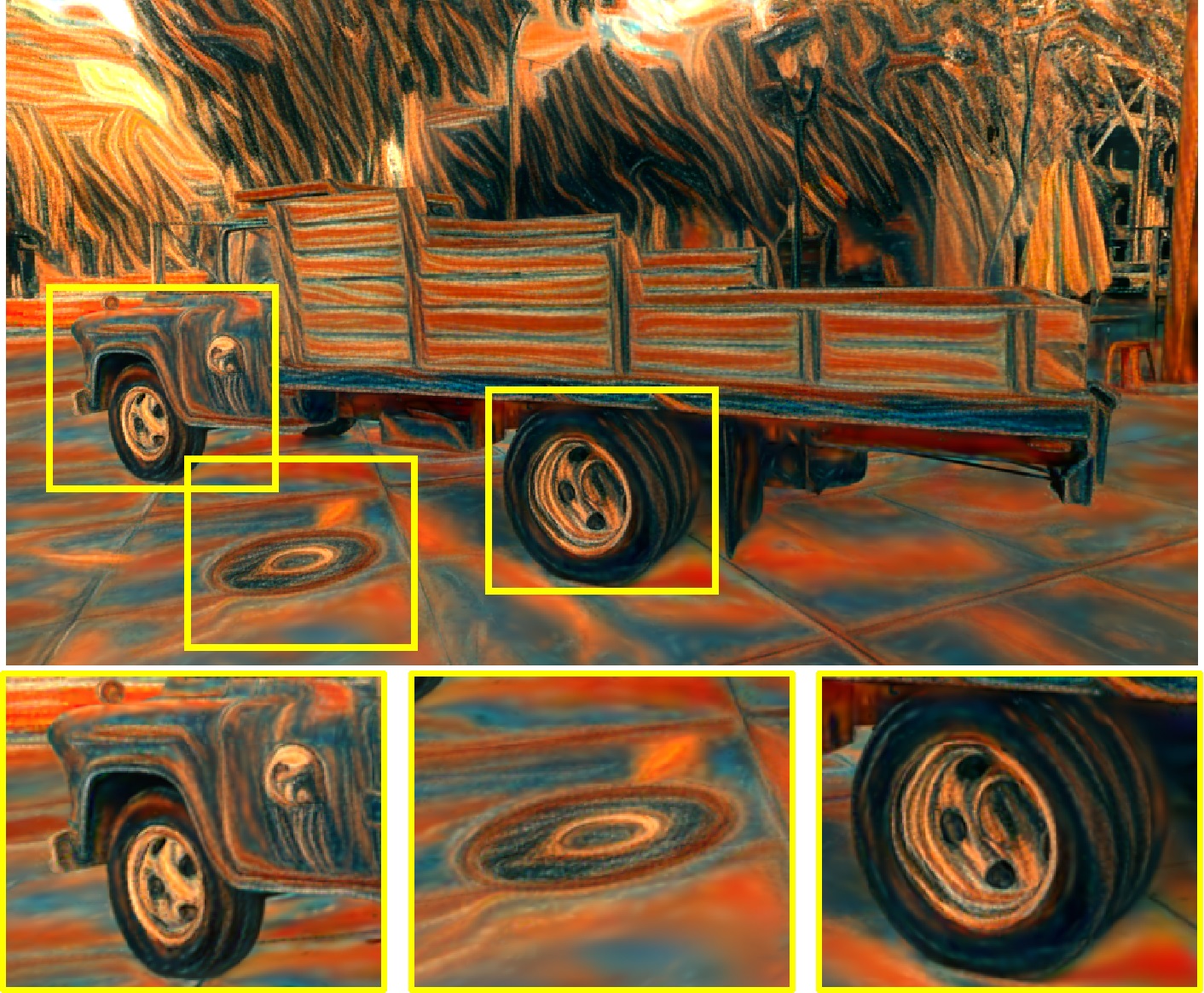}
    \caption{\textbf{Optimizing color only}} 
    \label{fig:sub1}
  \end{subfigure}
  \hspace{-5pt}
  \begin{subfigure}{0.49\textwidth} 
    \centering
    \includegraphics[width=\textwidth]{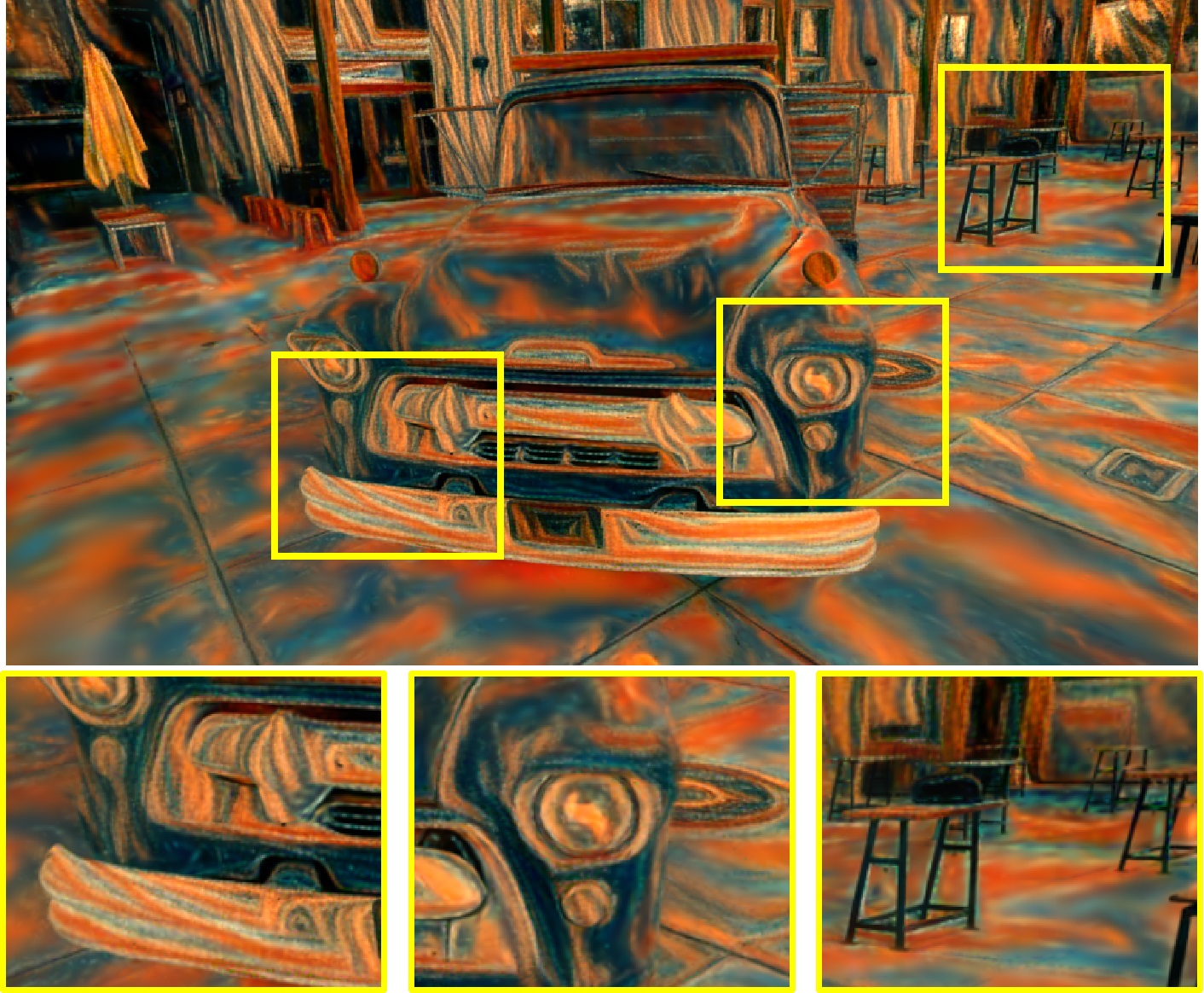}
    \caption{\textbf{Optimizing color only}} 
    \label{fig:sub1}
  \end{subfigure}
  \hspace{-5pt}
  \begin{subfigure}{0.49\textwidth} 
    \centering
    \includegraphics[width=\textwidth]{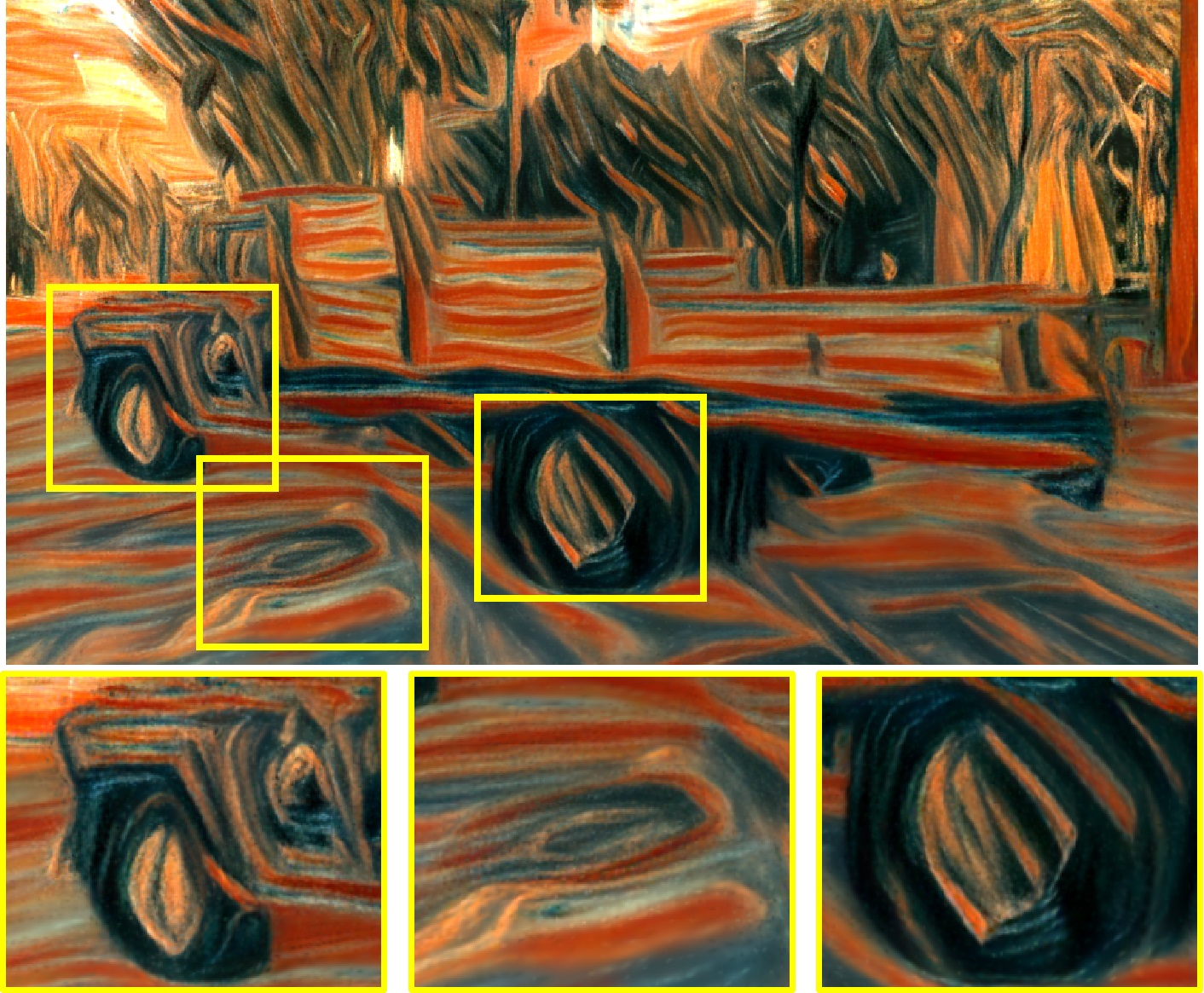}
    \caption{\textbf{Optimizing all paras}} 
    \label{fig:sub1}
  \end{subfigure}
  \hspace{-5pt}
  \begin{subfigure}{0.49\textwidth} 
    \centering
    \includegraphics[width=\textwidth]{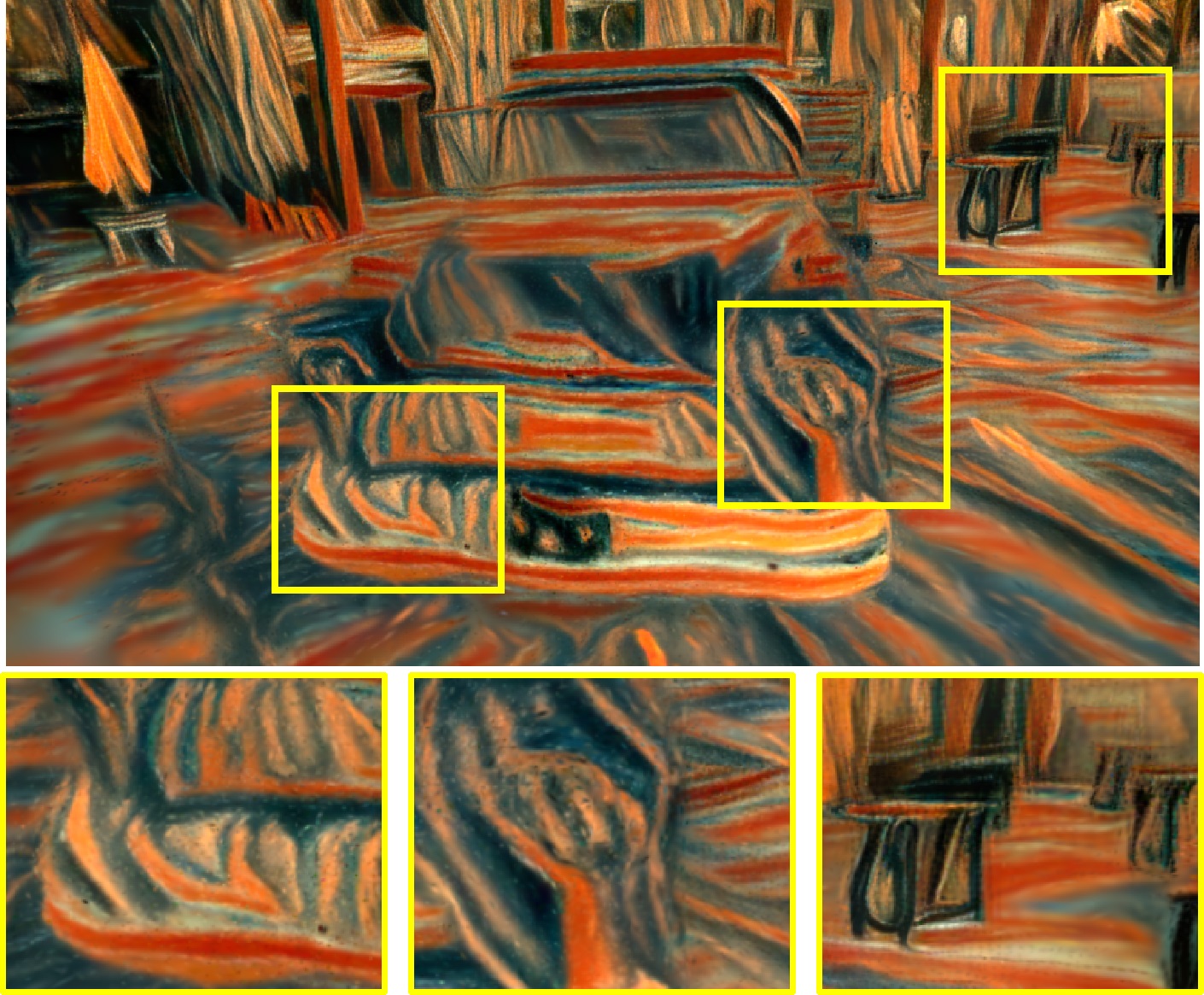}
    \caption{\textbf{Optimizing all paras}} 
    \label{fig:sub1}
  \end{subfigure}
  \hspace{-5pt}
  \caption{\textbf{Ablation study of optimizing color only.}}
  \label{fig:opt_geo_ablation}
\end{figure}

\subsubsection{Optimizing color only.}
During the reconstruction stage, all parameters of the 3D Gaussians must typically be optimized to accurately fit the real views. However, the primary objective of the stylization task is to transfer the artistic texture and color distribution of a reference image to the 3D scene while rigorously preserving the original geometric structure. If we continue optimizing all parameters during the stylization stage, the style loss will forcibly alter the spatial distribution and shape of the Gaussians to accommodate the reference image, which inevitably leads to severe collapse and artifacts in the underlying 3D geometry. Therefore, we propose freezing the geometric attributes of the 3D Gaussians and optimizing only the color parameters during stylization. To demonstrate this, we conducted specific ablation studies. By observing Fig.~\ref{fig:opt_geo_ablation}, it is evident that optimizing all parameters severely degrades the geometric structure of the scene. For instance, the truck's wheels lose their original circular contour and become heavily distorted, the front bumper blurs and exhibits melting-like visual artifacts, and the manhole cover on the ground are almost entirely lost. In contrast, optimizing color parameters successfully transfers the target artistic style to the scene surfaces while perfectly preserving the rigid geometric features of the original scene. The quantitative results presented in Tab.~\ref{tab:opt_geo_ablation} further confirm our viewpoint. Compared to optimizing all parameters, optimizing only the color parameters achieves significant improvements across all evaluation metrics. Most notably, the structure loss~\cite{splicing} drops substantially from 0.0607 to 0.0318, which objectively proves the superior capability of our strategy in maintaining the original 3D geometric fidelity of the scene. Furthermore, the ArtFID~\cite{artfid} metric, which measures the overall quality of stylization, improves from 23.215 to 22.801, indicating that color optimization can more stably fit the target style distribution. Regarding multi-view consistency, the Short and Long MEt3R~\cite{met3r} metrics decrease to 0.1196 and 0.2795, respectively. This suggests that by strictly anchoring the 3D geometry, the rendered images from novel viewpoints maintain better physical consistency.

\begin{table}[t]
\centering
\caption{Ablation study for optimizing color only.}
\label{tab:opt_geo_ablation}
\footnotesize
\setlength{\tabcolsep}{2.5pt} 
\begin{tabular}{lccccc}
\toprule


Optimized paras & ArtFID$\downarrow$ & Structure Loss$\downarrow$ & Short MEt3R$\downarrow$ & Long MEt3R$\downarrow$ \\

\midrule
All paras & 23.215 & 0.0607 & 0.1358 & 0.2905 \\
Color only & \textbf{22.801} & \textbf{0.0318} & \textbf{0.1196} & \textbf{0.2795} \\

\bottomrule
\end{tabular}
\end{table}

\begin{figure}
\captionsetup[subfigure]{labelformat=empty, labelsep=none}
  \centering
  \begin{subfigure}{0.49\textwidth} 
    \centering
    \includegraphics[height=18cm]{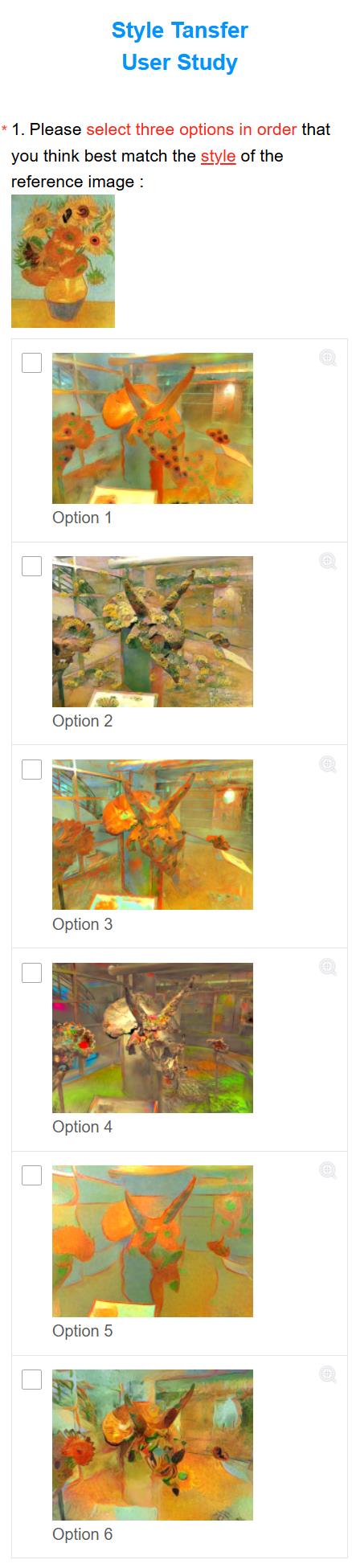}
    \caption{} 
    \label{fig:sub1}
  \end{subfigure}
  \hspace{-5pt}
  \begin{subfigure}{0.49\textwidth} 
    \centering
    \includegraphics[height=18cm]{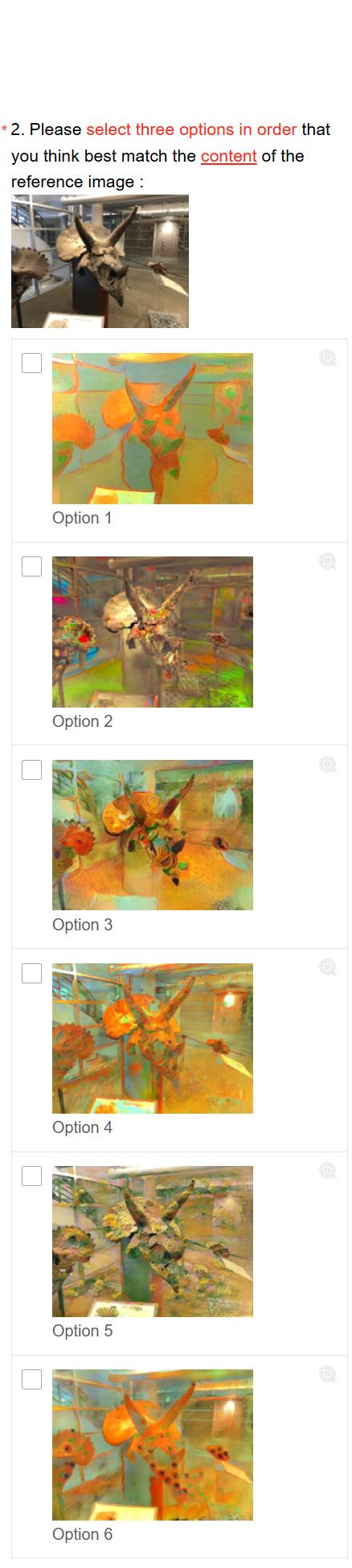}
    \caption{}
    \label{fig:sub2}
  \end{subfigure}
  \hspace{-5pt}
  \caption{\textbf{Interface of User Study.}}
  \label{fig:user_ui}
\end{figure}

\begin{figure}
\captionsetup[subfigure]{labelformat=empty, labelsep=none}
  \centering
  \begin{subfigure}{0.49\textwidth} 
    \centering
    \includegraphics[height=18cm]{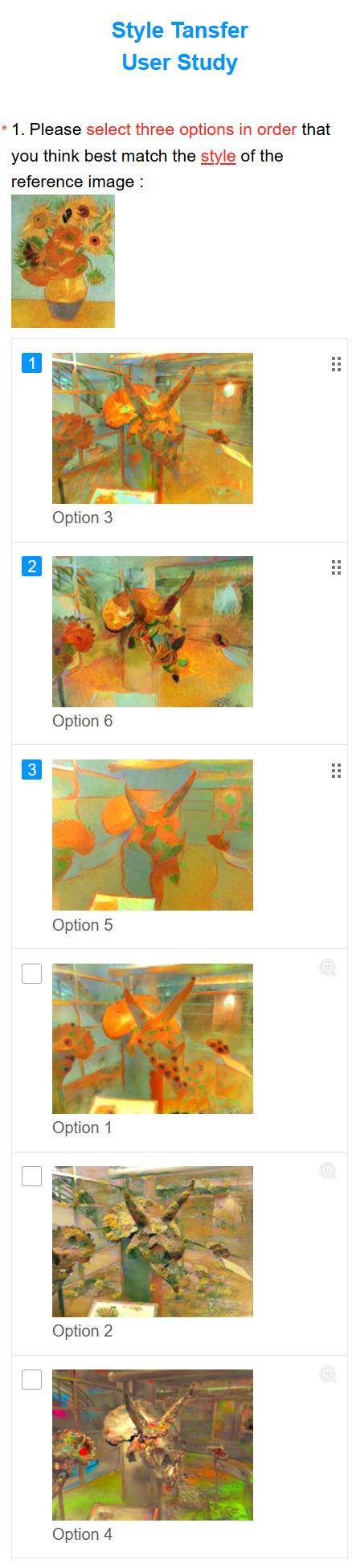}
    \caption{} 
    \label{fig:sub1}
  \end{subfigure}
  \hspace{-5pt}
  \begin{subfigure}{0.49\textwidth} 
    \centering
    \includegraphics[height=18cm]{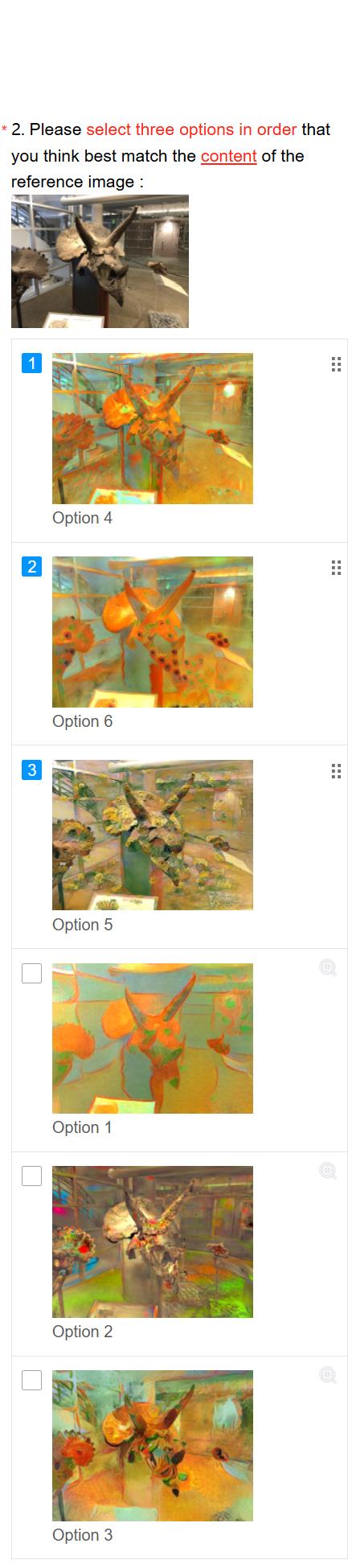}
    \caption{}
    \label{fig:sub2}
  \end{subfigure}
  \hspace{-5pt}
  \caption{\textbf{An example of user answers.}}
  \label{fig:answer_ui}
\end{figure}

\subsection{Additional Details of User Study}
\label{sec:user_study}

\subsubsection{Interface of user study.}
Fig.~\ref{fig:user_ui} illustrates the system interface employed for our user study. In this interface, each question presents a reference image along with six randomly ordered options. Each option displays the stylized result of one method. Participants are tasked with selecting three options in order from the option list that most closely match the reference image.

\subsubsection{An example of user answers.}
Fig.~\ref{fig:answer_ui} shows an example of user answers. Once a participant sequentially selects three options, the system automatically assigns numerical labels and reorders the selections. A label of "1" denotes the user’s primary choice—representing the highest perceived degree of similarity to the reference—while labels "2" and "3" are assigned to subsequent selections in descending order of preference. The methods in each rank will be assigned scores of 5 for the best, 3 for the second, 2 for the third, and 1 for the remaining.

\subsection{Additional Details of Our Feature Transport}
\label{sec:algorithm_details}

Our Capacity-Controlled Feature Transport (CCFT) Loss is based on existing optimal transport theories. The theory of optimal transport originates from the Monge problem~\cite{monge_ot}, which seeks an optimal transport map to rearrange a source mass distribution into a target distribution while satisfying the push-forward constraint and minimizing the total cost of transportation. However, the Monge problem is often ill-posed due to its highly non-linear nature and the fact that it does not allow for "mass splitting"—a map cannot send a single source point to multiple target destinations. To address these limitations, Kantorovich~\cite{klv_ot}proposed a fundamental relaxation of the Monge problem, which no longer requires the mass to be determined by a single mapping, but allows the mass of one source point to be allocated to multiple target points, thereby relaxing the original problem into a convex optimization form that is easier to analyze and solve. The discrete Kantorovich problem takes the following form:

\begin{equation}
\begin{gathered}
\min_{T \ge 0}\; \langle T, C\rangle = \sum_{(i,j)} T_{ij} C_{ij} \\
\text{s.t.} \quad T\mathbf{1}=a, \quad T^{\top}\mathbf{1}=b
\end{gathered}
\end{equation}

Here, $C \in \mathbb{R}_{+}^{n \times m}$ is the cost matrix, where $C_{ij}$ represents the cost of moving a unit mass from index $i$ to index $j$, and $T \in \mathbb{R}_{+}^{n \times m}$ is the transport matrix, where each entry $T_{ij}$ denotes the amount of mass transported from index $i$ to index $j$. The constraints ensure that the row and column sums of $T$ match the prescribed marginal distributions $a$ and $b$, respectively. While the Kantorovich formulation provides a flexible framework for Optimal Transport, its practical application to large-scale datasets is often hindered by significant computational demands. To address this issue, Cuturi~\cite{cuturi2013sinkhorn} augments the objective function with an entropic regularization term, the transport plan $T$ balances the total transportation cost against the smoothness of the mapping, as measured by its entropy. The regularized optimization problem is formulated as follows:

\begin{equation}
\begin{gathered}
\min_{T \ge 0}\; \left[\langle T, C\rangle - \varepsilon H(T)\right] \\
\text{where}\quad H(T) = -\sum_{(i,j)} T_{ij} (\log T_{ij} - 1)
\end{gathered}
\end{equation}

where $\varepsilon$ is the regularization parameter, and the regularization term $H(T)$ is defined by the negative entropy. This regularization yields several transformative advantages: it ensures the uniqueness of the optimal transport plan, makes the resulting "Sinkhorn distance" fully differentiable with respect to the input marginals, and enables the use of the Sinkhorn-Knopp algorithm to efficiently solve the optimal transport plan:

\begin{equation}
\label{eq:ot_1}
T = \mathrm{Diag}(u) \:K\: \mathrm{Diag}(v)
\end{equation}

where $K = \exp(-C/\varepsilon)$ is the Gibbs kernel, and $u \in \mathbb{R}^n_{+}, v \in \mathbb{R}^m_{+}$ are unknown scaling factors (dual vectors) to be determined, which constrained by the marginal requirements $T\mathbf{1} = a$ and $T^{\top}\mathbf{1} = b$, respectively. Substituting the scaling form into the first marginal constraint yields:

\begin{equation}
\label{eq:ot_2}
T\mathbf{1} = \mathrm{Diag}(u)\,K\,v = u \odot (Kv)
\end{equation}

where $\odot$ denotes the element-wise product. Setting this equal to the source marginal $a$ gives the corresponding relation for $u$:

\begin{equation}
\label{eq:ot_3}
u \odot (Kv)=a \;\Rightarrow\; u=\frac{a}{Kv}
\end{equation}

Similarly, applying the second marginal constraint to the transpose of the transport plan, we obtain:

\begin{equation}
\label{eq:ot_4}
T^{\top}\mathbf{1} = \mathrm{Diag}(v)\,K^{\top}u = v \odot (K^{\top}u)
\end{equation}

Equating this to the target marginal $b$ yields the corresponding relation for $v$:

\begin{equation}
\label{eq:ot_5}
v \odot (K^{\top}u)=b \;\Rightarrow\; v=\frac{b}{K^{\top}u}
\end{equation}

The Sinkhorn-Knopp algorithm~\cite{cuturi2013sinkhorn} proceeds by alternating between these two updates, effectively performing a row and column normalization of the matrix $K$ until the marginal constraints are satisfied. For a given iteration $t$, the updates are defined as:

\begin{equation}
\label{eq:ot_5}
u^{(t+1)}=\frac{a}{Kv^{(t)}},\quad
v^{(t+1)}=\frac{b}{K^{\top}u^{(t+1)}}
\end{equation}

After completing the iteration, the optimal transport plan $T^{*}$ is obtained by using the converged scaling vectors $u^{*}$ and $v^{*}$:

\begin{equation}
\label{eq:ot_6}
T^{*} = \mathrm{Diag}(u^{*}) \:K\: \mathrm{Diag}(v^{*})
\end{equation}

While the standard Sinkhorn algorithm effectively computes balanced transport, its rigid mass-conservation constraints limit its applicability in scenarios where the source and target distributions have unequal total mass or contain noise. To relax these constraints, Chizat et al.~\cite{uot} introduced a generalized framework for unblanced optimal transport, substituting the hard marginal constraints with KL-divergence, which extend the classical Sinkhorn distance to  the mass conservation is only enforced approximately. Inspired by this, we adopt a semi-balanced variant in our style transfer task, in which the source (content) marginal is preserved exactly, while the target (style) marginal is only softly constrained via KL-divergence. This allows only the matching style features to be transported and enables fine-grained control over the style pattern through the degree of constraint. Therefore, our semi-balanced optimal transport problem is formulated as:

\begin{equation}
\begin{gathered}
\min_{T \ge 0}\; \left[\langle T, C\rangle
-\varepsilon H(T)
+\tau\,\mathrm{KL}(T^{\top}\mathbf{1}\|b)\right] \\
\text{where}\quad \mathrm{KL}(x \,\|\, y)=\sum_k \left(x_k \log \frac{x_k}{y_k}-x_k+y_k\right)
\end{gathered}
\end{equation}

Here, the first-order optimality conditions for this objective still permit a diagonal scaling solution, we can still use Eq.\eqref{eq:ot_5} and Eq.\eqref{eq:ot_6} teratively solving the optimal transport plan $T$. The difference arises in the column update. Since the column marginal is no longer required to match $b$ exactly, the update for $v$ is obtained by balancing the entropy term against the KL-divergence. The first-order optimality condition gives:

\begin{equation}
\varepsilon \log v + \tau \log\!\left(\frac{T^{\top}\mathbf{1}}{b}\right) = 0,
\end{equation}

Substituting $T^{\top}\mathbf{1} = v \odot (K^{\top}u)$ and rearranging:

\begin{equation}
(\tau+\varepsilon)\log v = \tau \log\!\left(\frac{b}{(K^\top u)}\right),
\end{equation}

Therefore:

\begin{equation}
v^{(t+1)}=\left(\frac{b}{K^{\top}u^{(t+1)}}\right)^{\frac{\tau}{\tau+\varepsilon}}
\end{equation}

This update can be interpreted as a softened column projection. When $\tau \to \infty$, the update recovers the standard balanced Sinkhorn iteration, whereas a finite $\tau$ allows the transport plan to adaptively re-scale the mass to accommodate the cost of the transport $C$. This semi-unbalanced approach provides the necessary flexibility for style migration tasks, ensuring that the source structure is maintained while the target mass is utilized selectively. We describe the calculation process of our CCFT Loss in Algorithm~\ref{alg:topk_uot_guide_loss}.

\begin{algorithm}
\caption{Capacity-Controlled Feature Transport Loss}
\label{alg:topk_uot_guide_loss}
\DontPrintSemicolon

\KwIn{
Content feature $\{F_c\}$, style feature list $\{F_s\}$, 
Top-$K$ parameter $K$, entropy regularization $\varepsilon$, column marginal constraint $\tau$, Sinkhorn iterations $ST$, guide weight $\lambda_{guide}$.
}
\KwOut{Style loss $\mathcal{L}_{CCFT}$.}
\KwData{Guide map $G$.}

\BlankLine
$X \leftarrow \mathrm{L2Normalize}\!\left(\mathrm{FlattenSpatial}({F}_c)\right)$ \tcp*{$X \in \mathbb{R}^{n\times C}$}
$Y \leftarrow \mathrm{L2Normalize}\!\left(\mathrm{FlattenSpatial}({F}_s)\right)$ \tcp*{$Y \in \mathbb{R}^{m\times C}$}
$n \leftarrow \mathrm{rows}(X), \quad m \leftarrow \mathrm{rows}(Y)$\;

\BlankLine
$S \leftarrow X (Y + \lambda_{guide} G)^\top$\;

$(V_{\mathrm{topk}}, I_{\mathrm{topk}}) \leftarrow \mathrm{TopK}(S, K)$

$V^{\mathrm{style}}_{\mathrm{topk}} \leftarrow \mathrm{RowDot}\!\left(X,\; Y[I_{\mathrm{topk}}]\right)$\; $C_{\mathrm{topk}} \leftarrow \max(1 - V^{\mathrm{style}}_{\mathrm{topk}},\, 0)$\;

\BlankLine
$K_{\mathrm{sparse}} \leftarrow \exp\!\left(-\,\mathrm{stopgrad}(C_{\mathrm{topk}})/\varepsilon\right)$\;

$a_i \leftarrow \frac{1}{n}, \quad \forall i=1,\dots,n$\;
$\mathrm{deg}(j) \leftarrow \#\{(i,k): I_{\mathrm{topk}}[i,k]=j\}$\;
$\mathcal{J} \leftarrow \{j \mid \mathrm{deg}(j) > 0\}$\;
\For{$j=1$ \KwTo $m$}{
    $b_j \leftarrow
    \begin{cases}
        \frac{1}{|\mathcal{J}|}, & j \in \mathcal{J},\\
        10^{-12}, & \text{otherwise}
    \end{cases}$\;
}

$\alpha_{\mathrm{col}} = \frac{\tau}{\tau+\varepsilon}$\;

$u \leftarrow \mathbf{1}_n,\quad v \leftarrow \mathbf{1}_m$\;

\BlankLine
\For{$iter=1$ \KwTo $ST$}{
    $K v \leftarrow \mathrm{RowSum}\!\big(K_{\mathrm{sparse}} \odot v[I_{\mathrm{topk}}]\big)$\;
    $u \leftarrow a \oslash (K v)$\;
    $\mathrm{col\_sum} \leftarrow \mathbf{0}_m$\;
    \ForEach{$(i,k)$ in sparse Top-$K$ edges}{
        $\mathrm{col\_sum}[I_{\mathrm{topk}}[i,k]]
        \leftarrow
        \mathrm{col\_sum}[I_{\mathrm{topk}}[i,k]]
        + u_i \, K_{\mathrm{sparse}}[i,k]$\;
    }
    $v \leftarrow \big(b \oslash \mathrm{col\_sum}\big)^{\alpha_{\mathrm{col}}}$\;
}

\BlankLine
$T_{\mathrm{sparse}} \leftarrow \big(u \mathbf{1}_K^\top\big) \odot K_{\mathrm{sparse}} \odot v[I_{\mathrm{topk}}]$\;
$\mathcal{L}_{CCFT} \leftarrow \sum_{i=1}^{n}\sum_{k=1}^{K} T_{\mathrm{sparse}}[i,k]\; C_{\mathrm{topk}}[i,k]$\;

\BlankLine
$X2Y \leftarrow \mathbf{0}_{m\times C}$\;
\ForEach{$(i,k)$ in sparse Top-$K$ edges}{
    $j \leftarrow I_{\mathrm{topk}}[i,k]$\;
    $X2Y[j] \leftarrow X2Y[j] + T_{\mathrm{sparse}}[i,k] \, X[i]$\;
}
$G \leftarrow \mathrm{detach}\!\left(\mathrm{L2Normalize}(X2Y)\right)$\;

\Return{$\mathcal{L}_{CCFT}$}\;
\end{algorithm}

\end{document}